\def\year{2022}\relax
\documentclass[letterpaper]{article} % DO NOT CHANGE THIS
\usepackage{aaai22}  % DO NOT CHANGE THIS
\usepackage{times}  % DO NOT CHANGE THIS
\usepackage{helvet}  % DO NOT CHANGE THIS
\usepackage{courier}  % DO NOT CHANGE THIS
\usepackage[hyphens]{url}  % DO NOT CHANGE THIS
\usepackage{graphicx} % DO NOT CHANGE THIS
\usepackage{subfigure}
\usepackage{overpic}
\usepackage{natbib}  % DO NOT CHANGE THIS AND DO NOT ADD ANY OPTIONS TO IT
\usepackage{caption} % DO NOT CHANGE THIS AND DO NOT ADD ANY OPTIONS TO IT
\DeclareCaptionStyle{ruled}{labelfont=normalfont,labelsep=colon,strut=off} % DO NOT CHANGE THIS
\usepackage{algorithm}
\usepackage{algorithmic}
\usepackage{color}
\usepackage[table,xcdraw]{xcolor}
\usepackage{booktabs}  
\usepackage{multirow}
\usepackage{verbatim}
\usepackage{amsfonts}
\usepackage{amsmath}
\usepackage{mathrsfs}
\usepackage{amssymb}
\usepackage{authblk}

\newcommand{\ie}{\textit{i}.\textit{e}.}

\usepackage{makecell}
\usepackage[colorlinks,
            linkcolor=red,
            urlcolor=blue,
            anchorcolor=red,
            citecolor=black]{hyperref}
\usepackage{url}

\usepackage{newfloat}
\usepackage{listings}
\floatstyle{ruled}
\newfloat{listing}{tb}{lst}{}
\floatname{listing}{Listing}
\title{Pale Transformer: A General Vision Transformer Backbone with Pale-Shaped Attention}
\author{
    Sitong Wu\textsuperscript{\rm 1, \rm 2} 
    Tianyi Wu\textsuperscript{\rm 1, \rm 2}, 
    Haoru Tan\textsuperscript{\rm 3}, 
    Guodong Guo\textsuperscript{\rm 1, \rm 2}
    \thanks{Corresponding author.}
}
\affiliations{
    \textsuperscript{\rm 1}Institute of Deep Learning, Baidu Research, Beijing, China\\
	\textsuperscript{\rm 2}National Engineering Laboratory for Deep Learning Technology and Application, Beijing, China\\
	\textsuperscript{\rm 3}School of Artificial Intelligence, University of Chinese Academy of Sciences, Beijing, China\\
	
	\{wusitong, wutianyi01, guoguodong01\}@baidu.com,
	tanhaoru2018@ia.ac.cn
}

\usepackage{bibentry}
\begin{document}

\maketitle

% ========================= Abstract =========================

\begin{abstract}

Recently, Transformers have shown promising performance in various vision tasks. 
To reduce the quadratic computation complexity caused by the global self-attention, various methods constrain the range of attention within a local region to improve its efficiency.
%, and struggled for bridging the gap between local and global receptive fields.
Consequently, their receptive fields in a single attention layer are not large enough, resulting in insufficient context modeling.
To address this issue, we propose a Pale-Shaped self-Attention (PS-Attention), which performs self-attention within a pale-shaped region. 
Compared to the global self-attention, PS-Attention can reduce the computation and memory costs significantly. 
Meanwhile, it can capture richer contextual information under the similar computation complexity with previous local self-attention mechanisms. 
Based on the PS-Attention, we develop a general Vision Transformer backbone with a hierarchical architecture, named Pale Transformer, which achieves 83.4\%, 84.3\%, and 84.9\% Top-1 accuracy with the model size of 22M, 48M, and 85M respectively for $224\times224$ ImageNet-1K classification, outperforming the previous Vision Transformer backbones. For downstream tasks, our Pale Transformer backbone performs better than the recent state-of-the-art CSWin Transformer by a large margin on ADE20K semantic segmentation and COCO object detection \& instance segmentation. The code will be released on \url{https://github.com/BR-IDL/PaddleViT}. 

\end{abstract}

%in an efficient parallel way. 
%with UperNet and Mask R-CNN framework, respectively. 

% ========================= Introduction =========================

\section{Introduction}

\noindent Inspired by the success of Transformer \cite{Transformer} on a wide range of tasks in natural language processing (NLP) \cite{machine_translation, text_classification}, Vision Transformer (ViT) \cite{ViT} first employed a pure Transformer architecture for image classification, which shows the promising performance of Transformer architecture for vision tasks.
However, the quadratic complexity of the global self-attention results in expensive computation costs and memory usage especially for high-resolution scenarios, making it unaffordable for applications in various vision tasks.

\begin{figure*}[t]
	\centering
	% Global Self-Attention
 	\begin{minipage}{0.2\linewidth}
   		\centering
   		\subfigure[Global Self-Attention]{
			\hspace{0.14cm}
			\label{figure1a}\includegraphics[width=0.8\linewidth]{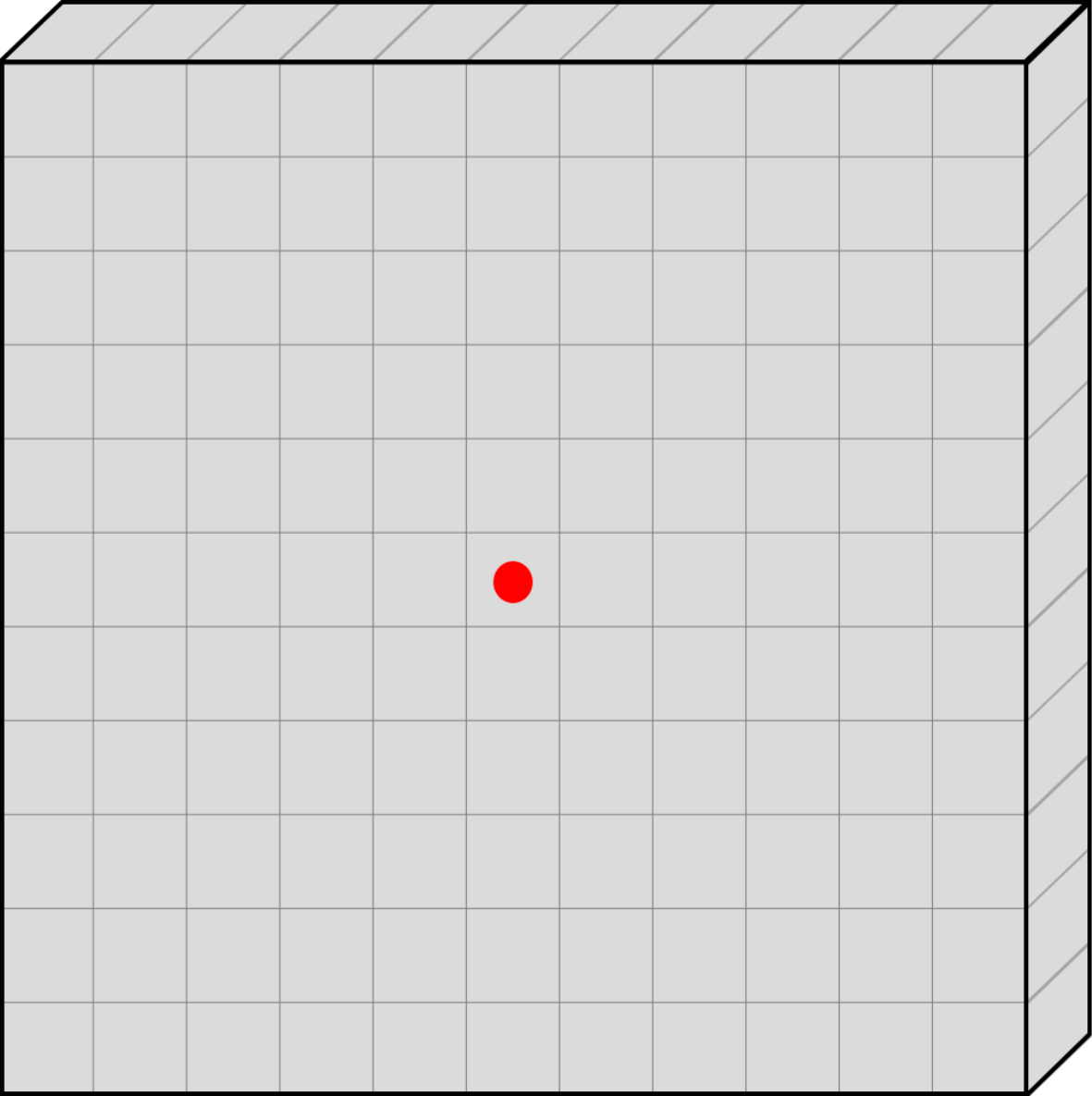}
			\hspace{0.06cm}
			}
  	\end{minipage}
 	%\hspace{0.05cm}
  	\begin{minipage}{0.78\linewidth}
   		\centering
		% Window based Self-Attention
		\subfigure[Window-based Self-Attention]{
			\label{figure1b}
			% Regular Window
			\begin{minipage}{0.2\linewidth}
				\centerline{\includegraphics[width=1.0\linewidth]{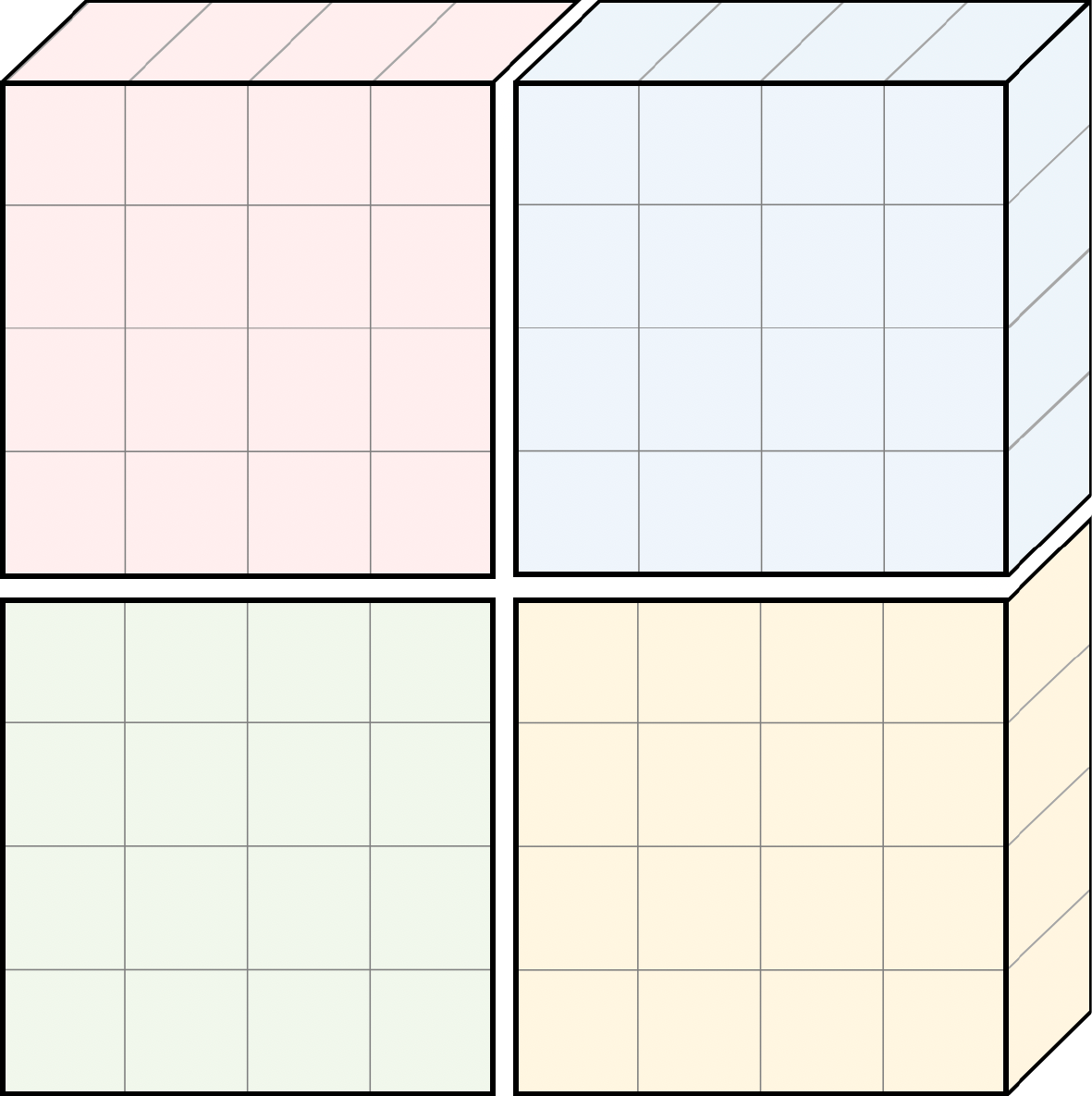}}
				\centerline{\scriptsize Regular Window}
				\vspace{0.2cm}
			\end{minipage}
    			\hspace{0.5cm}
			% Shifted Window
    			\begin{minipage}{0.2\linewidth}
				\centerline{\includegraphics[width=1.0\linewidth]{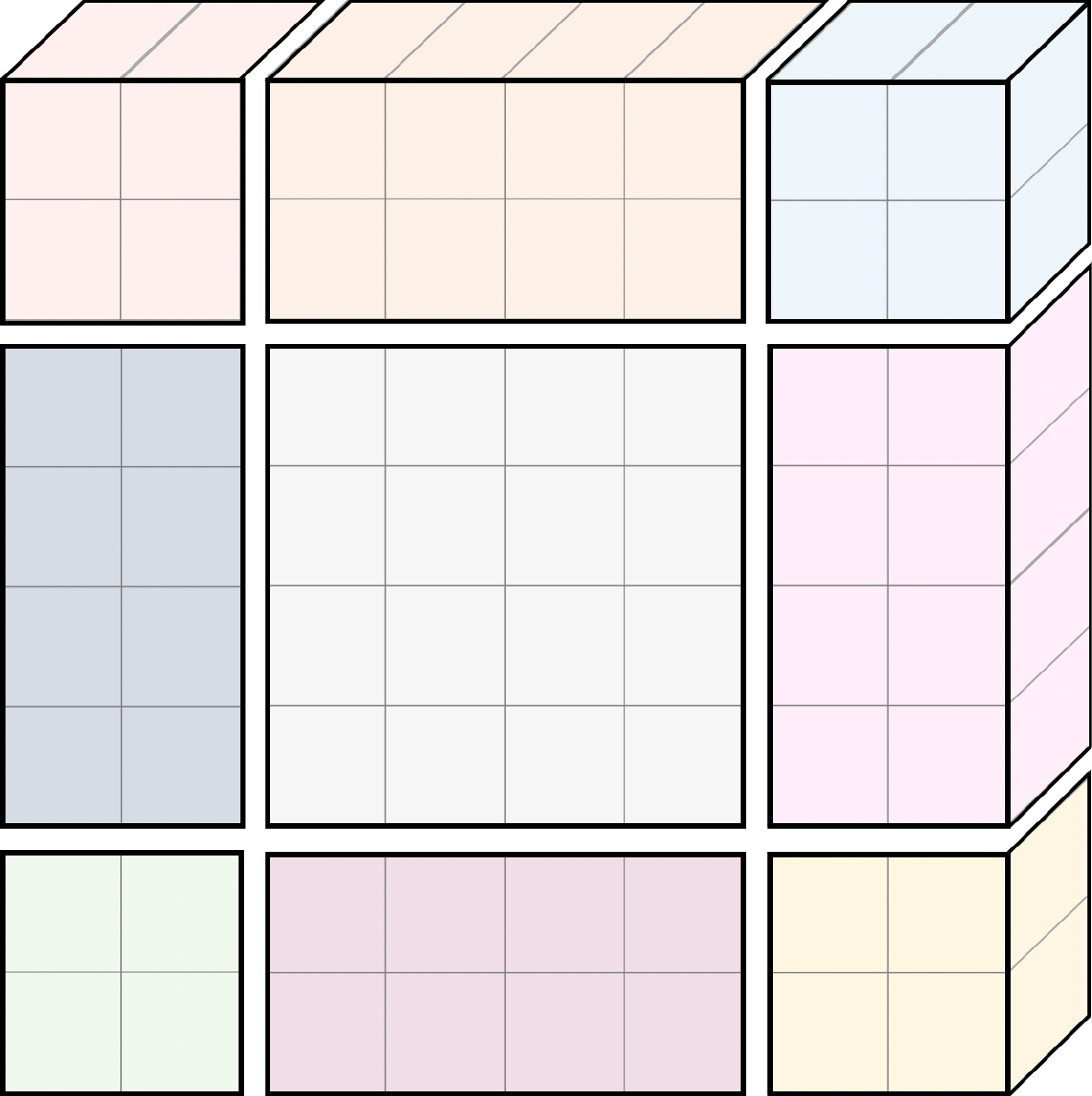}}
				\centerline{\scriptsize Shifted Window}
				\vspace{0.2cm}
			\end{minipage}
			\hspace{0.5cm}
			% Shuffle Window
			\begin{minipage}{0.2\linewidth}
				\centerline{\includegraphics[width=1.0\linewidth]{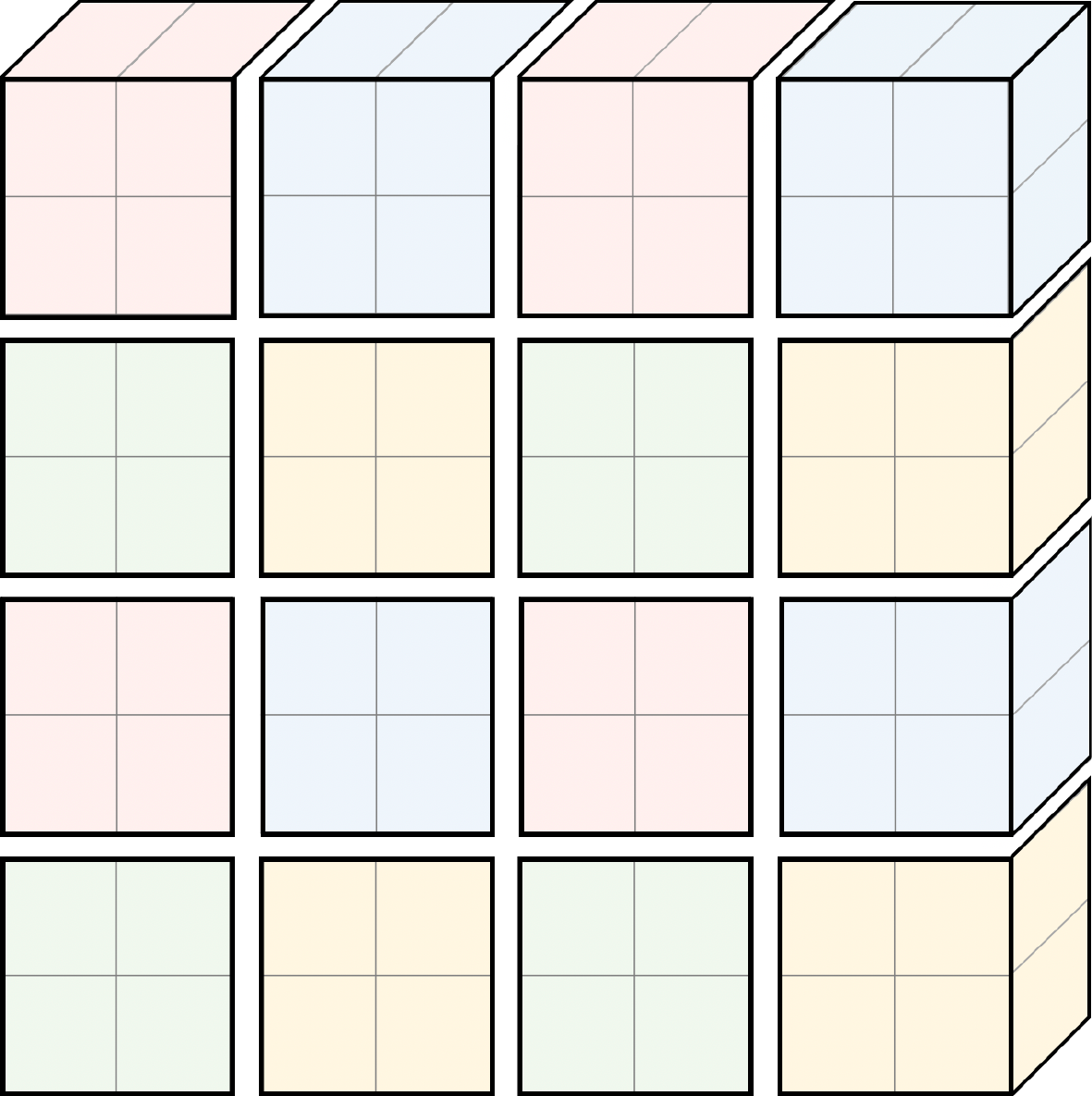}}
				\centerline{\scriptsize Shuffled Window}
				\vspace{0.2cm}
			\end{minipage}
			\hspace{0.5cm}
			%  Messenger
			\begin{minipage}{0.2\linewidth}
				\centerline{\includegraphics[width=1.0\linewidth]{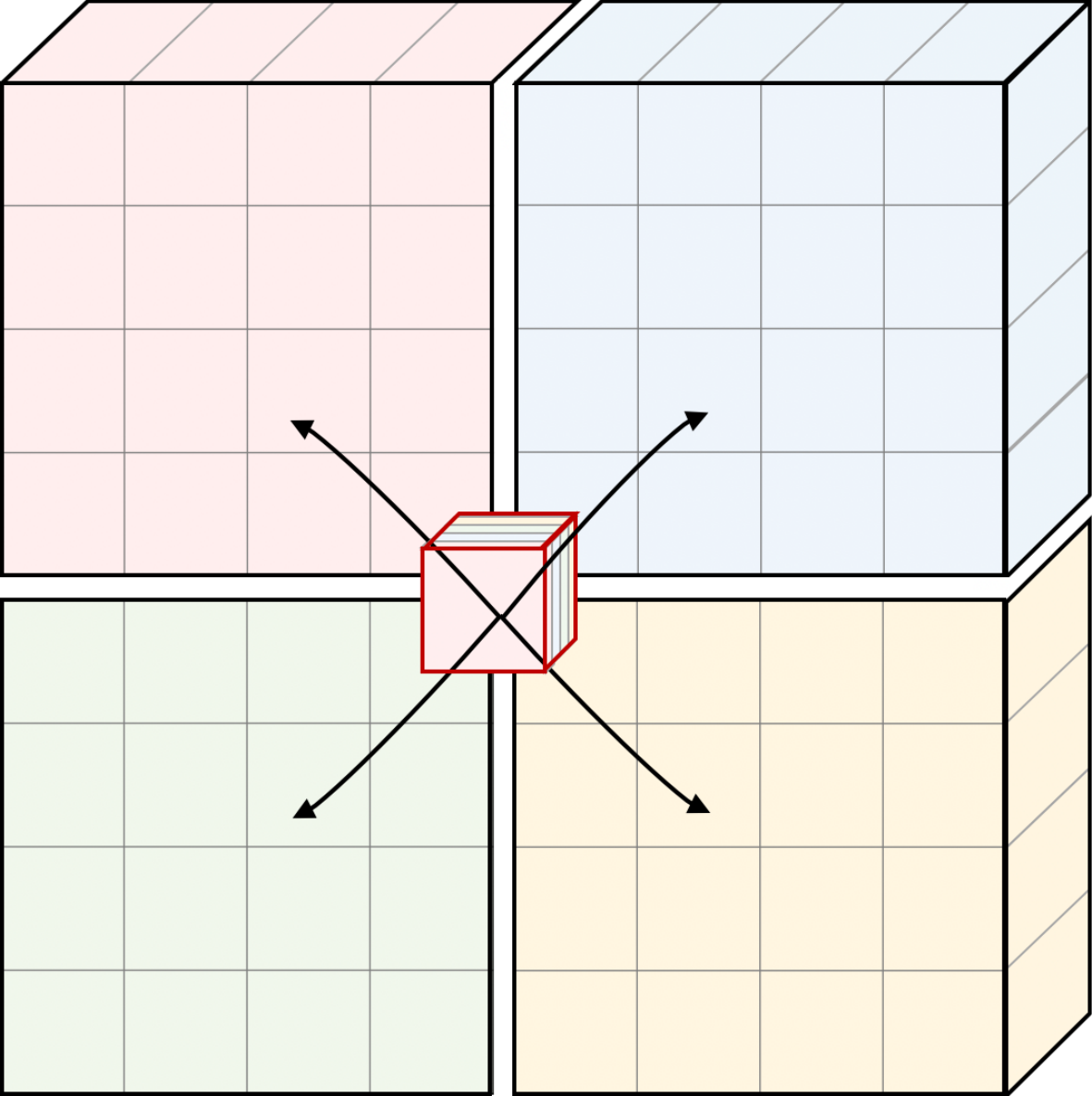}}
				\centerline{\scriptsize Messenger}
				\vspace{0.2cm}
			\end{minipage}
		}
		\subfigure{
			% axial 1
			\begin{minipage}{0.2\linewidth}
				\centerline{\includegraphics[width=1.0\linewidth]{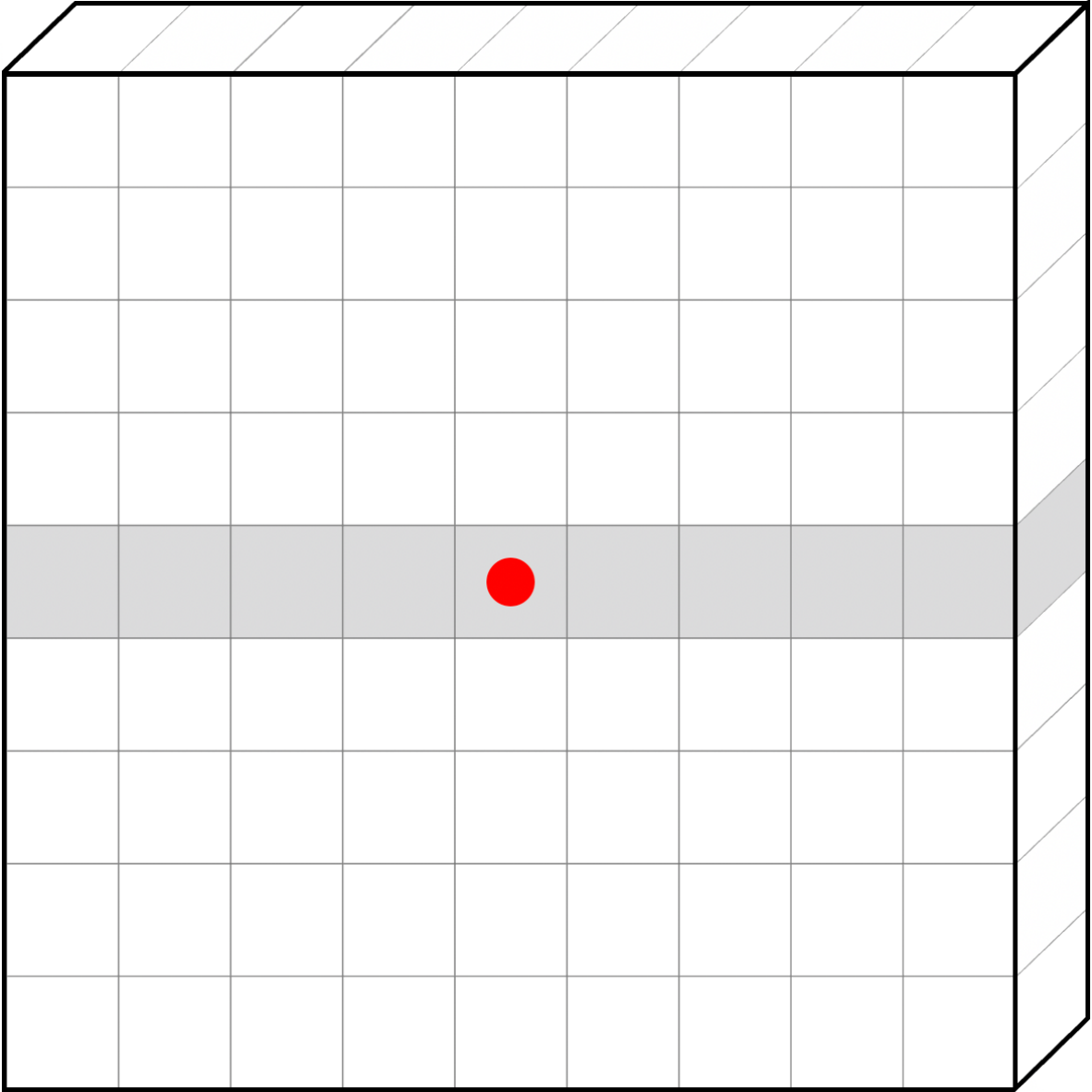}}
			\end{minipage}
    			\hspace{0.5cm}
			% axial 2
    			\begin{minipage}{0.2\linewidth}
				\centerline{\includegraphics[width=1.0\linewidth]{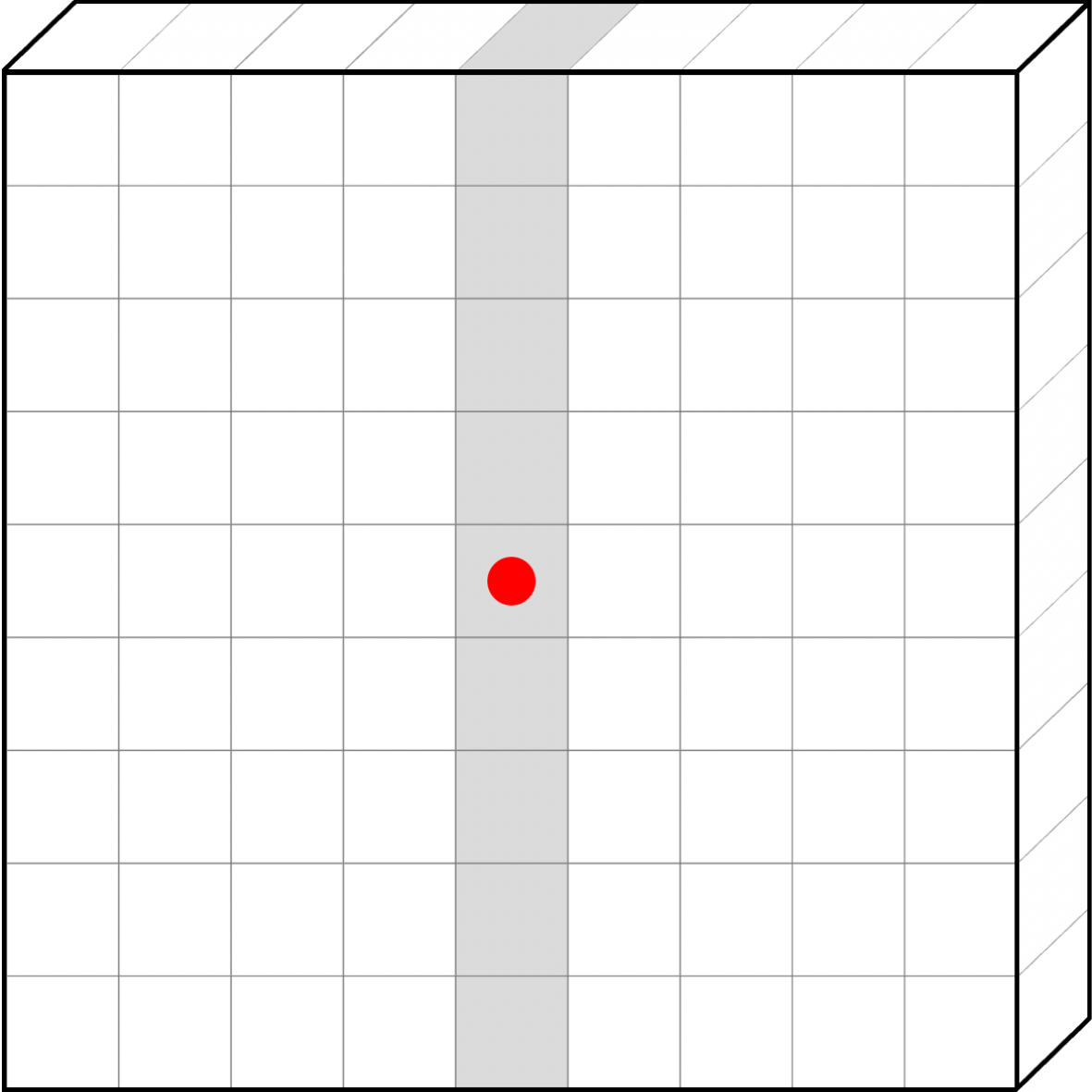}}
			\end{minipage}
			\hspace{0.5cm}
			% cswin
			\begin{minipage}{0.2\linewidth}
				\centerline{\includegraphics[width=1.0\linewidth]{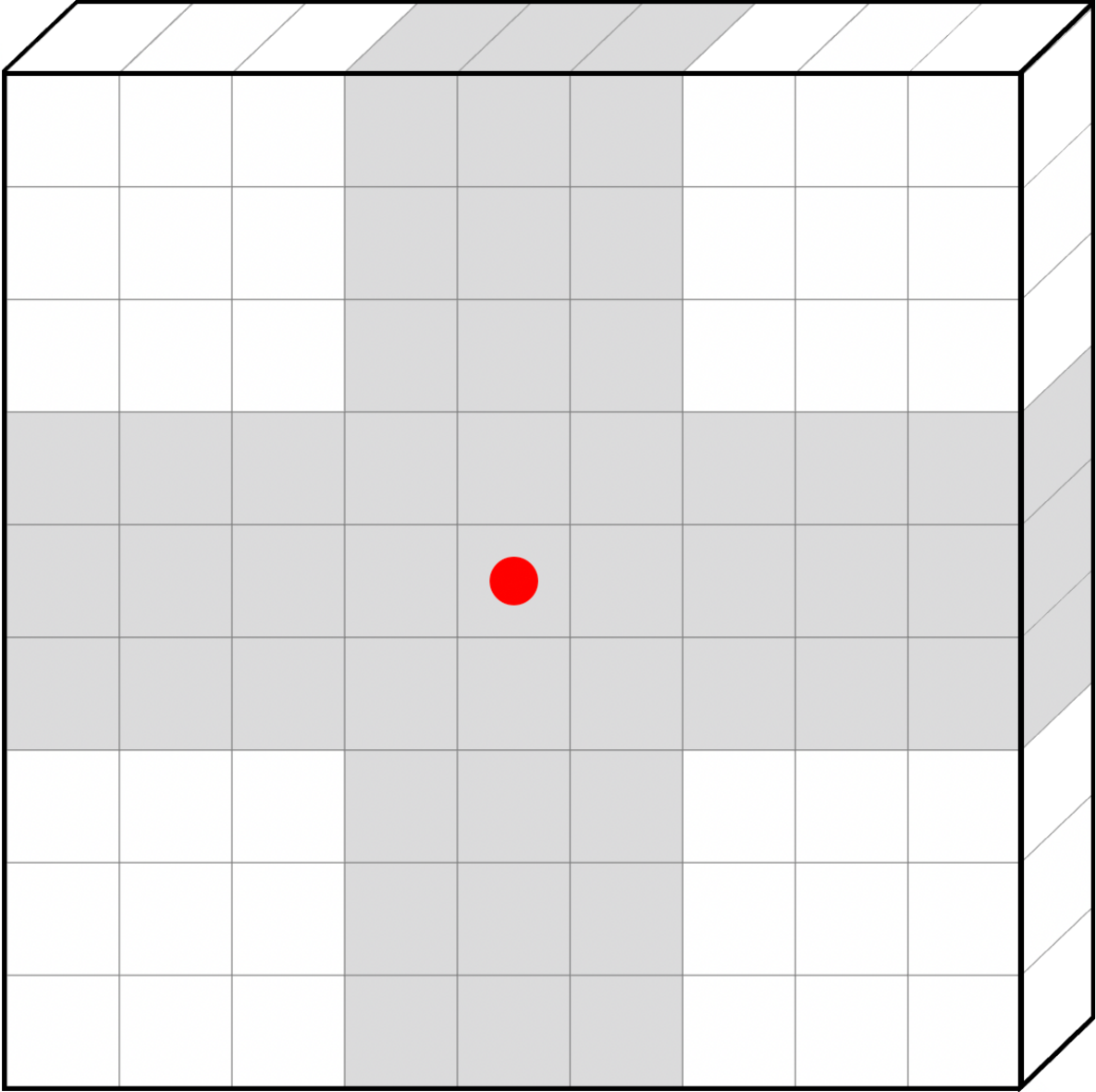}}
			\end{minipage}
			\hspace{0.5cm}
			%  pale
			\begin{minipage}{0.2\linewidth}
				\centerline{\includegraphics[width=1.0\linewidth]{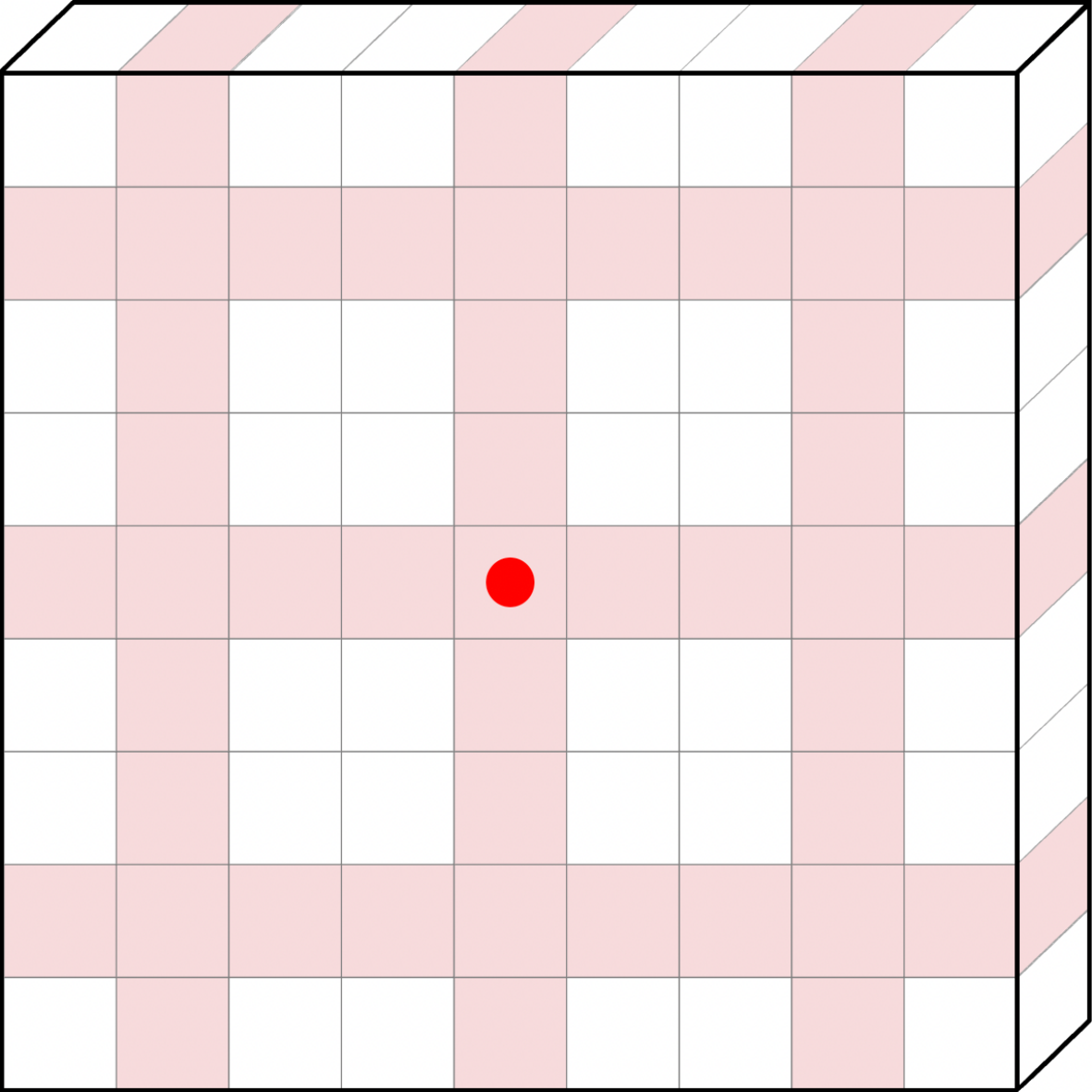}}
			\end{minipage}
		}
	\end{minipage}	
	\rightline{\small (c)  Axial Self-Attention \qquad \qquad \qquad \qquad \small (d) Cross-Shaped \quad \quad \quad \quad \small (e) Pale-Shaped \quad \quad \quad }
	\rightline{\small Window Self-Attention \ \ \quad  \ \ \small Self-Attention (ours) \quad \ \ }
	%\vspace{-0.45cm} 
	%\rightline{\qquad \qquad \qquad \qquad \qquad \qquad \qquad \qquad \qquad \qquad \ \ \ }
	\caption{\label{figure1}Illustration of different self-attention mechanisms in Transformer backbones. (a) is the standard global self-attention. (b) Window-based self-attention mechanisms perform attention inside each window, and introduce various strategies to build cross-window connections. Different colors in (b) represent different windows. In (c), (d), and (e), the input features are first split into multiple groups, one of which is illustrated by the shadow area, and the self-attention is conducted within each group. Thus, for a reference token denoted by the red dot, it can interact directly with the tokens covered by the shadow area.}
\end{figure*}

A typical way to improve the efficiency is to replace the global self-attention with local ones. 
A crucial and challenging issue is how to enhance the modeling capability under the local settings. 
% 列举之前的工作：Swin、Shuffle、MSG、Axial attention、CSwin
For example, Swin \cite{Swin} and Shuffle Transformer \cite{Shuffle} proposed shifted window and shuffled window, respectively (Figure {\ref{figure1b}}), and alternately used two different window partitions (i.e., regular window and the proposed window) in consecutive blocks to build cross-window connections. MSG Transformer \cite{MSG} manipulated the messenger tokens to exchange information across windows. Axial self-attention \cite{Axial_Deeplab} treated the local attention region as a single row or column of the feature map (Figure {\ref{figure1}\textcolor{red}{(c)}}).
%, which provides long-range dependencies in horizontal and vertical directions. 
CSWin \cite{CSwin} proposed cross-shaped window self-attention (Figure {\ref{figure1}\textcolor{red}{(d)}}), which can be regarded as a multiple row and column expansion of axial self-attention. 
% 尽管这些方法取得了不错的结果，但是它们在single attention layer的range不够丰富，以至于不能提取充分的上下文关系。
Although these methods achieve excellent performance and are even superior to the CNN counterparts, the dependencies in each self-attention layer are not rich enough for capturing sufficient contextual information.

% --------------------- Figure 1. Attention Comparison  ---------------------

In this work, we propose a Pale-Shaped self-Attention (PS-Attention) to capture richer contextual dependencies efficiently. Specifically, the input feature maps are first split into multiple pale-shaped regions spatially. 
Each pale-shaped region (abbreviating as pale) is composed of the same number of interlaced rows and columns of the feature map.
The intervals between adjacent rows or columns are equal for all the pales.
For example, the pink shadow in Figure {\ref{figure1}\textcolor{red}{(e)}} indicates one of the pales. 
Then, self-attention is performed within each pale. For any token, 
it can directly interact with other tokens within the same pale, which endows our method with the capacity of capturing richer contextual information in a single PS-Attention layer. To further improve the efficiency, we develop a more efficient parallel implementation of the PS-Attention. Benefit from the larger receptive fields and stronger context modeling capability, our PS-Attention shows superiority to the existing local self-attention mechanisms illustrated in Figure \ref{figure1}.

% Obviously, our PS-Attention has stronger context modeling and representation learning capacity than the existing local self-attention mechanisms (as illustrated in Figure \ref{figure1}). 

% benefited from the richer contextual dependencies                      dependencies varied from short-range to long-range 

Based on the proposed PS-Attention, we design a general vision transformer backbone with a hierarchical architecture, named Pale Transformer.
We scale our approach up to get a series of models, including Pale-T (22M), Pale-S (48M), and Pale-B (85M),
reaching significantly better performance than previous approaches.
Our Pale-T achieves 83.4\% Top-1 classification accuracy on ImageNet-1k, 50.4\% single-scale mIoU on ADE20K (semantic segmentation), 47.4 box mAP (object detection) and 42.7 mask mAP (instance segmentation) on COCO, outperforming the state-of-the-art backbones by +0.7\%, +1.1\%, +0.7, and +0.5, respectively. 
Furthermore, our largest variant Pale-B is also superior to the previous methods, achieving 84.9\% Top-1 accuracy on ImageNet-1K, 52.2\% single-scale mIoU on ADE20K, 49.3 box mAP and 44.2 mask mAP on COCO. 

%To demonstrate the effectiveness of our Pale Transformer, we conduct extensive experiments on ImageNet-1K classification and various downstream tasks. 

% ======================== Related works =========================

\section{Related Work}

%CNN以前一直是视觉任务的主流架构，这种趋势被ViT打破，它使用纯transformer的架构来做图像分类（将图像视作patch序列）。尽管在训练数据充足的情况下，ViT可以达到于CNNs相匹敌或者更好的分类性能，但是它仍然存在很多局限性有待解决。自此，越来越多的transformer backbones涌现来改进原始的ViT，它们主要focus on两个方面：（1）如何增强视觉transformer的locality（2）如何达到性能和效率之间的平衡
ViT \cite{ViT}, which takes the input image as a sequence of patches, has paved a new way and shown promising performance for many vision tasks dominated by CNNs over the years. 
A line of previous Vision Transformer backbones mainly focused on the following two aspects to better adapt to vision tasks: (1) Enhancing the locality of Vision Transformers. (2) Seeking a better trade-off between performance and efficiency.

% ----- 1 -----
\subsection{Locally-Enhanced Vision Transformers} 
% 与CNN不同的是，transformer 没有局部连接这种归纳偏置，这可能会导致 local structure 的提取不够，例如线、边缘和颜色。很多工作致力于加强 Vision Transformers 的局部特征建模能力。
Different from CNNs, the inductive bias for local connections is not involved in the original Transformer, which may lead to insufficient extraction of local structures, such as lines, edges, and color conjunctions. Many works are devoted to strengthening the local feature extraction of Vision Transformers. 
% 最早的方法是：将ViT的单一尺度改为层次化架构来构建多尺度特征。% 另一个思路是: 结合CNN和transformer，例如结合方式有两种：
%（1）Introduce conv module into transformer，例如 Local VIT / CvT / CMT
%（2）通过 CNN & transformer 双分支的结构，来结合双方的优势: Mobile-Former / Conformer / DS-Net
% 此外，一些方法通过 integrating local attention 来获得更细粒度的特征。TNT / Twins / ViL / LG Transformer / VOLO
The earliest approach is to replace the single-scale architecture of ViT with a hierarchical one to obtain multi-scale features \cite{PVT}. 
Such design is followed by many works afterward \cite{Swin,Shuffle,Focal,CSwin}.
Another way is to combine CNNs and Transformers. Mobile-Former \cite{Mobile-Former}, Conformer \cite{Conformer} and DS-Net \cite{DS-Net} integrated the CNN and Transformer features by the well-designed dual-branch structures. In contrast, Local ViT \cite{LocalViT}, CvT \cite{CvT} and Shuffle Transformer \cite{Shuffle} only inserted several convolutions into some components of Transformer.
Besides, some works obtain richer features by fusing the multi-branch with different scales \cite{CrossViT} or cooperating with local attention \cite{TNT,ViL,Twins,LG_Transformer,VOLO}.

%, such as patch merging layer \cite{CvT}, multi-head self-attention layer \cite{CvT}, feed forward network \cite{LocalViT}, after the multi-head self-attention layer \cite{Shuffle} and before the first patch merging layer. 

% --------------------- Figure 2. Overall Architecture  ---------------------
\begin{figure*}[t]
\centering
\includegraphics[width=0.95 \linewidth]{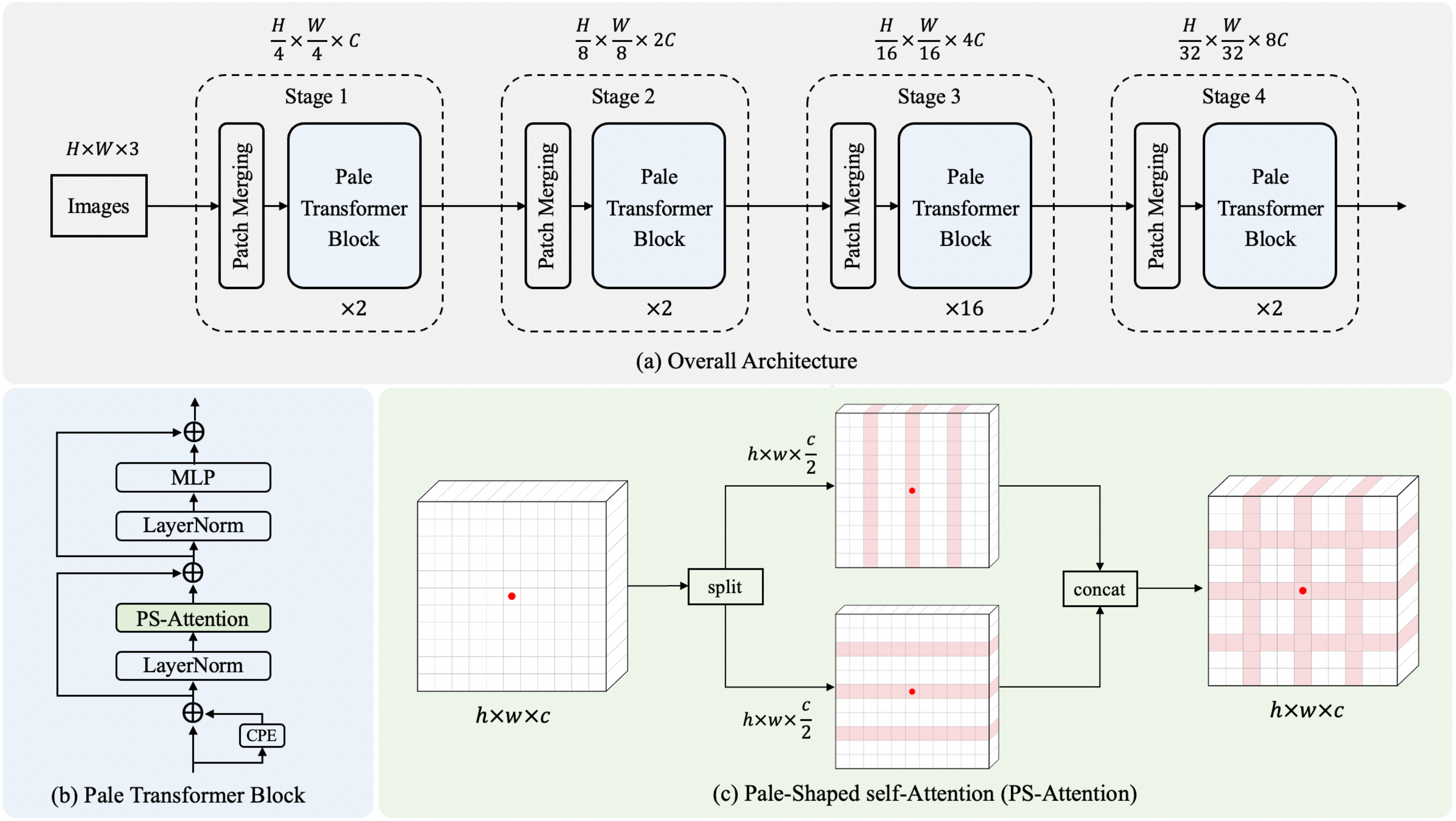}
\caption{(a) The overall architecture of our Pale Transformer. (b) The composition of each block. (c) Illustration of parallel implementation of PS-Attention. For a reference token (red dot), it can directly interact with the tokens within the shadow area.}
\label{figure2}
\end{figure*}

% ----- 2 -----
\subsection{Efficient Vision Transformers}
% 标准的Vision Transformer的计算复杂度与输入图像尺寸成二次方的关系，这对于高分辨率的输入是无法承受的。提高效率的方法有两种：（1）通过剪枝策略减少冗余计算（2）设计更有效的self-attention机制。
%The computation complexity of ViT \cite{ViT} is quadratic to the input image size, which is unaffordable to high-resolution inputs in many applications. 
The mainstream research on improving the efficiency for Vision Transformer backbones has two folds: reducing the redundant calculations via pruning strategies and designing more efficient self-attention mechanisms.

\paragraph{\textbf{Pruning Strategies for Vision Transformers.}} 

For pruning, the existing methods can be divided into three categories: (1) Token Pruning. DVT \cite{DVT} proposed a cascade Transformer architecture to adaptively adjust the number of tokens according to the hardness for classification of the input image. Considering that tokens with irrelevant or even confusing information may be detrimental to image classification, some works proposed to locate discriminative regions and progressively drop less informative tokens by learnable sampling \cite{DynamicViT, PS-ViT} and reinforcement learning \cite{IA-RED2} strategies. However, such unstructured sparsity results in incompatibility with dense prediction tasks. Some structure-preserving token selection strategies were implemented via token pooling \cite{PSViT} and a slow-fast updating \cite{Evo-ViT}. (2) Channel Pruning. VTP \cite{VTP} presented a simple but effective framework to remove the reductant channels. (3) Attention Sharing. Based on the observation that attention maps from continuous blocks are highly correlated, PSViT \cite{PSViT} was proposed to reuse the attention calculation process between adjacent layers.

 % Another way is 
 
\paragraph{\textbf{Efficient Self-Attention Mechanisms.}} 
% 考虑到二次方的计算复杂度主要是self-attention引起的，很多工作都致力于在不损失性能的情况下提高attention的效率。
Considering that the quadratic computation complexity is caused by self-attention, many methods are committed to improving its efficiency while avoiding performance decay \cite{PVT,Deformable_DETR,Swin,Shuffle}. 
% ---------------  一种方式是降低K和V的数量 --------------- 
One way is to reduce the sequence length of key and value. PVT \cite{PVT} proposed a spatial reduction attention to downsample the scale of key and value before computing attention. Deformable attention \cite{Deformable_DETR} used a linear layer to select several keys from the full set, which can be regarded as a sparse version of global self-attention. However, excessive downsampling will lead to information confusion, and deformable attention relies heavily on a high-level feature map learned by CNN and may not be directly used on the original input image.
% --------------- 另一种方式是用local替换global --------------- 
Another way is to replace the global self-attention with local self-attention, which limits the range of each self-attention layer into a local region. As shown in Figure {\ref{figure1b}}, the feature maps are first divided into several non-overlapping square regular windows (indicated with diverse colors), and the self-attention is performed within each window individually. 
The key challenge for the design of local self-attention mechanisms is to bridge the gap between local and global receptive fields. 
A typical manner is to build connections across regular square windows. For example, alternately using regular window and another newly designed window partition manner (shifted window \cite{Swin} or shuffled window \cite{Shuffle} in Figure {\ref{figure1b}}) in consecutive blocks, and manipulating messenger tokens to exchange information across windows \cite{MSG}. Besides, axial self-attention \cite{Axial_Deeplab} achieves longer-range dependencies in horizontal and vertical directions respectively by performing self-attention in each single row or column of the feature map. CSWin \cite{CSwin} proposed a cross-shaped window self-attention region including multiple rows and columns. 
Although these existing local attention mechanisms can provide opportunities for breaking through the local receptive fields to some extent, their dependencies are not rich enough to capture sufficient contextual information in a single self-attention layer, which limits the modeling capacity of the whole network.

% ------------------------ 点明我们的goal，和我们的方法的优点 ------------------------ 
%Therefore, we are committed to designing an effective and efficient self-attention mechanism, Pale-Shaped self-Attention (PS-Attention).
%It can cover richer contextual dependencies inside a single attention layer than the previous local attention mechanisms, which contributes to stronger context modeling capacity.

The most related to our work is CSWin \cite{CSwin}, which developed a cross-shaped window self-attention mechanism for computing self-attention in the horizontal and vertical stripes, while our proposed PS-Attention computes self-attention in the pale-shaped regions. 
Moreover, the receptive fields of each token in our method are much wider than CSWin, which also endows our approach with stronger context modeling capacity.

%\begin{figure*}[t]
%\centering
%\includegraphics[width=0.8\textwidth]{figure2} % Reduce the figure size so that it is slightly narrower than the column.
%\caption{Adjusting the bounding box instead of actually removing the unwanted data resulted multiple layers in this paper. It also needlessly increased the PDF size. In this case, the size of the unwanted layer doubled the paper's size, and produced the following surprising results in final production. Crop your figures properly in a graphics program. Don't just alter the bounding box.}
%\label{fig2}
%\end{figure*}

% ========================== Method ===========================
\section{Methodology} 
In this section, we first present our Pale-Shaped self-Attention (PS-Attention) and its efficient parallel implementation.
%followed by complexity analysis to compare the theoretical computation costs. 
%Then, the composition of each Pale Transformer block is given. 
Then, the composition of the Pale Transformer block is given. 
Finally, we describe the overall architecture and variants configurations of our Pale Transformer backbone.

\subsection{Pale-Shaped self-Attention}

For capturing dependencies varied from short-range to long-range, 
we propose Pale-Shaped self-Attention (PS-Attention), 
which computes self-attention within a pale-shaped region (abbreviating as pale). 
As shown in the pink shadow of Figure \ref{figure1}\textcolor{red}{(e)}, 
one pale contains $s_r$ interlaced rows and $s_c$ interlaced columns, which covers a region containing $(s_rw + s_ch - s_rs_c)$ tokens. We define $(s_r, s_c)$ as the pale size. 
Given an input feature map $X\in\mathcal{R}^{h \times w \times c}$, 
we first split it into multiple pales $\{P_1, ..., P_N\}$ with the same size $(s_r, s_c)$,
where $P_i \in\mathcal{R}^{(s_rw + s_ch - s_rs_c) \times c}, i\in\{1, 2, ... ,N\}$. 
The number of pales is equal to $N=\frac{h}{s_r}=\frac{w}{s_c}$, 
which can be ensured by padding or interpolation operation.
% for integral pale division. 
For all pales, intervals between adjacent rows or columns are the same. The self-attention is then performed within each pale individually.
As illustrated in Figure \ref{figure1}, the receptive field of PS-Attention is significantly wider and richer than all the previous local self-attention mechanisms, enabling more powerful context modeling capacity.

%Inspired by atrous convolution \cite{chen2017deeplab} and axial attention \cite{Axial_Deeplab}, 
%we design a pale-shaped attention region to build richer dependencies ranges for contextual modeling. 
%First of all, we introduce the concept of pale. As shown in the pink shadow of Figure \ref{figure1}(e), a pale-shaped region (referred to as a pale for short) contains $s_r$ interlaced rows and $s_c$ interlaced columns, which covers a region containing $(s_rw + s_ch - s_rs_c)$ tokens. We define $(s_r, s_c)$ as the pale size. 

%Given an input feature map $X\in\mathcal{R}^{h \times w \times c}$, we first split it into multiple pales $\{P_1, ..., P_N\}$ with the same size $(s_r, s_c)$, where $P_i \in\mathcal{R}^{(s_rw + s_ch - s_rs_c) \times c}, i\in\{1, 2, ... ,N\}$. The number of pales follows the formula $N=\frac{h}{s_r}=\frac{w}{s_c}$, which can be ensured by padding or interpolation operation for integral pale division. For all pales, the interval between adjacent rows or columns is consistent. The self-attention is then performed within each pale respectively. As illustrated in Figure \ref{figure1}, the receptive field of PS-Attention is significantly wider and richer than the previous local-attention mechanisms, resulting in more powerful context modeling capacity.

\subsubsection{Efficient Parallel Implementation.}
%To further improve the efficiency, 
%we decompose the vanilla PS-Attention mentioned above into row-wise and column-wise attention, 
%after which the integrated features along these two orthogonal directions are merged together as the pale-wise attentive representations. 

%Specifically, as shown in Figure \ref{figure2}(c), we first divide the input feature $X\in\mathcal{R}^{h \times w \times c}$ into two independent parts $X_r\in\mathcal{R}^{h \times w \times \frac{c}{2}}$ and $X_c\in\mathcal{R}^{h \times w \times \frac{c}{2}}$ in the channel dimension, which are then split into multiple groups for row-wise and column-wise attention respectively.
%\begin{equation} 
%    \begin{aligned}
%    \label{pale:split}
%	& X_r = [X_r^1, ..., X_r^{N_r}], X_c = [X_c^1, ..., X_c^{N_c}],
%    \end{aligned}
%\end{equation}
%where $N_r = h/s_r$, $N_c = w/s_c$, $X_r^i \in \mathcal{R}^{s_r \times w \times c}$ contains $s_r$ interlaced rows, and $X_c^j  \in \mathcal{R}^{h \times s_c \times c}$ contains $s_c$ interlaced columns. 

To further improve the efficiency, we decompose the vanilla PS-Attention mentioned above into row-wise and column-wise attention, 
which perform self-attention within row-wise and column-wise token groups, respectively.
Specifically, as shown in Figure \ref{figure2}\textcolor{red}{(c)}, we first divide the input feature $X\in\mathcal{R}^{h \times w \times c}$ into two independent parts $X_r\in\mathcal{R}^{h \times w \times \frac{c}{2}}$ and $X_c\in\mathcal{R}^{h \times w \times \frac{c}{2}}$ in the channel dimension, which are then split into multiple groups for row-wise and column-wise attention respectively.
\begin{equation} 
    \begin{aligned}
    \label{pale:split}
	& X_r = [X_r^1, ..., X_r^{N_r}], X_c = [X_c^1, ..., X_c^{N_c}],
    \end{aligned}
\end{equation}
where $N_r = h/s_r$, $N_c = w/s_c$, $X_r^i \in \mathcal{R}^{s_r \times w \times c}$ contains $s_r$ interlaced rows, and $X_c^j  \in \mathcal{R}^{h \times s_c \times c}$ contains $s_c$ interlaced columns.

Then, the self-attention is conducted within each row-wise and column-wise token group, respectively. Similar to \cite{CvT}, we use three separable convolution layers $\phi_Q$, $\phi_K$, and $\phi_V$ to generate the query, key, and value. 
\begin{equation} 
    \begin{aligned}
    \label{pale:attention}
	& Y_r^i = {\rm MSA}(\phi_Q(X_r^i), \phi_K(X_r^i), \phi_V(X_r^i)), \\
	& Y_c^i = {\rm MSA}(\phi_Q(X_c^i), \phi_K(X_c^i), \phi_V(X_c^i)),  \\
    \end{aligned}
\end{equation}
where $i \in \{1, 2, ... ,N\}$, and MSA indicates the Multi-head Self-Attention \cite{ViT}.

Finally, the outputs of row-wise and column-wise attention are concatenated along channel dimension, 
resulting in the final output $Y\in\mathcal{R}^{h \times w \times c}$,
\begin{equation} 
    \begin{aligned}
    \label{pale:concat}
	& Y = {\rm Concat}(Y_r, Y_c),
    \end{aligned}
\end{equation}
where $Y_r = [Y_r^1, ..., Y_r^{N_r}]$ and $Y_c = [Y_c^1, ..., Y_c^{N_c}]$.

Compared to the vanilla implementation of PS-Attention within the whole pale, such a parallel mechanism has a lower computation complexity. 
Furthermore, the padding operation only needs to ensure $h$ can be divisible by $s_r$ and $w$ can be divisible by $s_c$, rather than $\frac{h}{s_r}=\frac{w}{s_c}$. Therefore, it is also conducive to avoiding excessive padding.

\subsubsection{Complexity Analysis.}
Given the input feature of size $h\times w\times c$ and pale size $(s_r, s_c)$, the standard global self-attention has a computational complexity of 
\begin{equation} 
    \mathcal{O}_\text{Global} = 4hwc^2 + 2c(hw)^2, 
    \label{eq_complex_global}
\end{equation}
however, our proposed PS-Attention under the parallel implementation has a computational complexity of 
\begin{equation} 
  \mathcal{O}_\text{Pale}  = 4hwc^2 + hwc(s_ch+s_rw+27) << \mathcal{O}_\text{Global},
    \label{eq_complex_pale_parallel}
\end{equation}
which can obviously alleviate the computation and memory burden compared with the global one, since $2hw >> (s_ch+s_rw+27)$ always holds. The detailed derivations of Eq. \eqref{eq_complex_global} and Eq. \eqref{eq_complex_pale_parallel} are provided in the supplementary material.

% Block
\subsection{Pale Transformer Block}
As shown in Figure \ref{figure2}\textcolor{red}{(b)}, our Pale Transformer block consists of three sequential parts, the conditional position encoding (CPE) for dynamically generating the positional embedding, the proposed PS-Attention module for capturing contextual information, and the MLP module for feature projection. The forward pass of the $l$-th block can be formulated as follows:
\begin{align}
& \tilde{X}^l = X^{l-1} + \text{CPE}({X^{l-1}})
 \label{eq_CPE},
 \\
& \hat{X}^l = \tilde{X}^l + \text{PS-Attention}\Big(\text{LN}(\tilde{X}^l)\Big) 
\label{eq:attention},
\\
& X^l = \hat{X}^l + \text{MLP\Big(LN}(\hat{X}^l)\Big),
\label{eq:mlp}
\end{align}
where LN($\cdot$) refers to layer normalization \cite{LN}. The CPE \cite{CPVT} is implemented as a simple depth-wise convolution, which is widely used in previous works \cite{FTN, Twins} for its compatibility with an arbitrary size of input. The PS-Attention module defined in Eq. \eqref{eq:attention} is constructed by sequentially performing Eq. \eqref{pale:split} to Eq. \eqref{pale:concat}. The MLP module defined in Eq. \eqref{eq:mlp} consists of two linear projection layers to expand and contract the embedding dimension sequentially, which is the same as \cite{ViT} for fair comparisons.

\subsection{Overall Architecture and Variants}
As illustrated in Figure \ref{figure2}\textcolor{red}{(a)}, the Pale Transformer consists of four hierarchical stages for capturing multi-scale features by following the popular design in CNNs \cite{ResNet} and Transformers \cite{Swin, CSwin}.
Each stage contains a patch merging layer and multiple Pale Transformer blocks. 
The patch merging layer aims to spatially downsample the input features by a certain ratio and expand the channel dimension by twice for a better representation capacity. For fair comparisons, we use the overlapping convolution for patch merging, the same as \cite{CvT, CSwin}. Specifically, the spatial downsampling ratio is set to 4 for the first stage and 2 for the last three stages, implementing by $7\times7$ convolution with stride 4 and $3\times3$ convolution with stride 2, respectively. The outputs of the patch merging layer are fed into the subsequent Pale Transformer blocks, with the number of tokens kept constant. Following \cite{Swin, CSwin}, we simply apply an average pooling operation on the top of the last block to obtain a representative token for the final classification head, which is composed of a single linear projection layer.

\subsubsection{Variants.}
The definitions of model hyper-parameters for the $i$-th stage are listed below:
\begin{itemize}
\item $P_i$: the spatial reduction factor for patch merging layer,
\item $C_i$: the embedding dimension of tokens,
\item $S_i$: the pale size for the PS-Attention,
\item $H_i$: the head number for the PS-Attention,
\item $R_i$: the expansion ratio for the MLP module.
\end{itemize}

By varying the hyper-parameters $H_i$ and $C_i$ in each stage, we design three variants of our Pale Transformer, named Pale-T (Tiny), Pale-S (Small), and Pale-B (Base), respectively. Table \ref{variants} shows the detailed configurations of all variants. 
Note that all variants have the same depth with $[2, 2, 16, 2]$ in four stages. In each stage of these variants, we set the pale size $s_r=s_c = S_i = 7$, and use the same MLP expansion ratio of $R_i = 4$.
Thus, the main differences among Pale-T, Pale-S, and Pale-B lie in the embedding dimension of tokens and the head number for the PS-Attention in four stages, \ie, variants vary from narrow to wide.

% ----------------- Table: Variants -----------------
\begin{table}[t]
\newcommand{\tabincell}[2]{\begin{tabular}{@{}#1@{}}#2\end{tabular}}
\renewcommand{\arraystretch}{1.0}
\centering
\resizebox{0.475\textwidth}{!}{
\begin{tabular}{ c | c | c | c | c | c }
\toprule[1pt]
\multirow{2}{*}{Stage}    & \multirow{2}{*}{\tabincell{c}{Output \\ Stride}}    & \multirow{2}{*}{Layer}    & \multirow{2}{*}{Pale-T}     & \multirow{2}{*}{Pale-S}      & \multirow{2}{*}{Pale-B}    \\
&  &    &   &       &    \\
\midrule[1pt]
\multirow{5.5}{*}{1}    & \multirow{5.5}{*}{4}    & \multirow{2}{*}{\tabincell{c}{Patch\\Merging}}     & \multirow{2}{*}{\tabincell{c}{$P_1=4$\\$C_1=64$}}     & \multirow{2}{*}{\tabincell{c}{$P_1=4$\\$C_1=96$}}     & \multirow{2}{*}{\tabincell{c}{$P_1=4$\\$C_1=128$}}   \\ 
&  &    &   &       &    \\
%\cline{3-6}
\cmidrule{3-6}
& &\multirow{3}{*}{\tabincell{c}{Pale \\Transformer \\ Block}}   & \multirow{3}{*}{$\begin{bmatrix}  S_1=7 \\ H_1=2 \\ R_1=4 \end{bmatrix} \times 2$}    &  \multirow{3}{*}{$\begin{bmatrix}  S_1=7 \\ H_1=2 \\ R_1=4 \end{bmatrix} \times 2$}     & \multirow{3}{*}{$\begin{bmatrix}  S_1=7 \\ H_1=4 \\ R_1=4 \end{bmatrix} \times 2$}   \\ 
&  &    &   &       &    \\
&  &    &   &       &    \\
\midrule[0.5pt]
\multirow{5.5}{*}{2}   & \multirow{5.5}{*}{8}        & \multirow{2}{*}{\tabincell{c}{Patch\\Merging}}     & \multirow{2}{*}{\tabincell{c}{$P_2=2$\\$C_2=128$}}     & \multirow{2}{*}{\tabincell{c}{$P_2=2$\\$C_2=192$}}     & \multirow{2}{*}{\tabincell{c}{$P_2=2$\\$C_2=256$}}   \\ %
&  &    &   &       &    \\
%\cline{3-6}
\cmidrule{3-6}
& & \multirow{3}{*}{\tabincell{c}{Pale \\Transformer \\ Block}}   & \multirow{3}{*}{$\begin{bmatrix}  S_2=7 \\ H_2=4 \\ R_2=4 \end{bmatrix} \times 2$}    & \multirow{3}{*}{$\begin{bmatrix}  S_2=7 \\ H_2=4 \\ R_2=4 \end{bmatrix} \times 2$}    & \multirow{3}{*}{$\begin{bmatrix}  S_2=7 \\ H_2=8 \\ R_2=4 \end{bmatrix} \times 2$}   \\ 
&  &    &   &       &    \\
&  &    &   &       &    \\
\midrule[0.5pt]          
\multirow{5.5}{*}{3}   & \multirow{5.5}{*}{16}        & \multirow{2}{*}{\tabincell{c}{Patch\\Merging}}   & \multirow{2}{*}{\tabincell{c}{$P_3=2$\\ $C_3=256$}}     & \multirow{2}{*}{\tabincell{c}{$P_3=2$\\$C_3=384$}}     & \multirow{2}{*}{\tabincell{c}{$P_3=2$\\$C_3=512$}}   \\ %
&  &    &   &       &    \\
%\cline{3-6}
\cmidrule{3-6}
& & \multirow{3}{*}{\tabincell{c}{Pale \\Transformer \\ Block}}   & \multirow{3}{*}{$\begin{bmatrix}  S_3=7 \\ H_3=8 \\ R_3=4 \end{bmatrix} \times 16$}    & \multirow{3}{*}{$\begin{bmatrix}  S_3=7 \\ H_3=8 \\ R_3=4 \end{bmatrix} \times 16$}  & \multirow{3}{*}{$\begin{bmatrix}  S_3=7 \\ H_3=16 \\ R_3=4 \end{bmatrix} \times 16$}   \\ 
&  &    &   &       &    \\
&  &    &   &       &    \\
\midrule[0.5pt]    
\multirow{5.5}{*}{4}   & \multirow{5.5}{*}{32}        & \multirow{2}{*}{\tabincell{c}{Patch\\Merging}}         & \multirow{2}{*}{\tabincell{c}{$P_4=2$\\$C_4=512$}}     & \multirow{2}{*}{\tabincell{c}{$P_4=2$\\$C_4=768$}}     & \multirow{2}{*}{\tabincell{c}{$P_4=2$\\$C_4=1024$}}   \\ %
&  &    &   &       &    \\
%\cline{3-6}
\cmidrule{3-6}
& & \multirow{3}{*}{\tabincell{c}{Pale \\ Transformer \\ Block}}   & \multirow{3}{*}{$\begin{bmatrix}  S_4=7 \\ H_4=16 \\ R_4=4 \end{bmatrix} \times 2$}    & \multirow{3}{*}{$\begin{bmatrix}  S_4=7 \\ H_4=16 \\ R_4=4 \end{bmatrix} \times 2$}  & \multirow{3}{*}{$\begin{bmatrix}  S_4=7 \\ H_4=32 \\ R_4=4 \end{bmatrix} \times 2$}   \\ 
&  &    &   &       &    \\
&  &    &   &       &    \\
\bottomrule[1pt]    
\end{tabular}}            
\caption{\label{variants} Detailed configurations of Pale Transformer Variants.}
\end{table}

% -------------------- ImageNet-1k --------------------
\begin{table}[!t]
\newcommand{\tabincell}[2]{\begin{tabular}{@{}#1@{}}#2\end{tabular}}
\centering
\resizebox{0.47\textwidth}{!}{
    \begin{tabular}{ l | cc | c }
    \toprule[1pt]
    \multirow{2}{*}{Backbone} & \multirow{2}{*}{Params} & \multirow{2}{*}{FLOPs} & \multirow{2}{*}{\tabincell{l}{Top-1 \\ \ \ (\%)}} \\ %\hline
    &      &     & \\
    \midrule[1pt] 
    %ResNet-50 \cite{ResNet}              & 26M      & 4.1G     & 78.5         \\
    RegNetY-4G \cite{RegNet}              & 21M       & 4.0G    & 80.0         \\ %\hline
    DeiT-S \cite{DeiT}                       & 22M      & 4.6G     & 79.8         \\
    PVT-S \cite{PVT}                        & 25M      & 3.8G     & 79.8         \\
    T2T-14 \cite{T2TViT}                   & 22M      & 6.1G      & 80.7         \\
    % LocalViT-S \cite{LocalViT}            & 22M      & 4.6G     & 80.8         \\
    DPT-S \cite{DPT}                        & 26M      & 4.0G     & 81.0         \\
    TNT-S \cite{TNT}                        & 24M      & 5.2G     & 81.3         \\
    Swin-T \cite{Swin}                          & 29M      & 4.5G     & 81.3         \\
    Twins-SVT-S \cite{Twins}            & 24M      & 2.8G     & 81.3         \\
    CvT-13 \cite{CvT}                        & 20M      & 4.5G     & 81.6         \\
    % PiT-S \cite{PiT}                           & 24M      & 2.9G.    & 81.9         \\
    ViL-S \cite{ViL}                           & 25M      & 4.9G     & 82.0         \\
    PVTv2-B2 \cite{PVTv2}                   & 25M      & 4.0G     & 82.0         \\
    Focal-T \cite{Focal}                        & 29M      & 4.9G     & 82.2         \\
    % CrossViT-15 \cite{CrossViT}             & 28M      & 6.1G      & 82.3         \\
    % RegionViT-S \cite{RegionViT}           & 31M      & 5.3G      & 82.5         \\
    % CrossFormer-S \cite{CrossFormer}   & 31M       & 4.9G      & 82.5         \\
    Shuffle-T \cite{Shuffle} & 29M      & 4.6G      & 82.5         \\
    CSWin-T \cite{CSwin}     & 23M      & 4.3G      & 82.7         \\
    LV-ViT-S$^{\star}$ \cite{LV-ViT}     & 26M      & 6.6G      & 83.3         \\
    \textbf{Pale-T} (ours)    & 22M      & 4.2G      & \textbf{83.4}         \\ %\hline
    \textbf{Pale-T$^{\star}$} (ours)    & 22M      & 4.2G      & \textbf{84.2}         \\ %\hline
    \midrule[0.5pt]
    %ResNet-101 \cite{ResNet}   & 45M       & 7.9G      & 79.8  \\
    RegNetY-8G \cite{RegNet}   & 39M       & 8.0G      & 81.7  \\ 
    PVT-M \cite{PVT}       & 44M       & 6.7G      & 81.2  \\
    T2T-19 \cite{T2TViT}    & 39M       & 9.8G      & 81.4  \\
    DPT-M \cite{DPT}    & 46M       & 6.9G      & 81.9  \\
    CvT-21 \cite{CvT}      & 32M       & 7.1G      & 82.5  \\
    % CrossViT-18 \cite{CrossViT}          & 44M       & 9.5G     & 82.8  \\
    Swin-S \cite{Swin}        & 50M       & 8.7G     & 83.0  \\
    MViT-B-24 \cite{MViT}                & 54M       & 10.9G    & 83.1  \\
    Twins-SVT-B \cite{Twins}            & 56M       & 8.3G     & 83.1  \\
    PVTv2-B3 \cite{PVTv2}   & 45M       & 6.9G     & 83.2  \\
    ViL-M \cite{ViL}        & 40M       & 8.7G      & 83.3  \\
    % RegionViT-M+ \cite{RegionViT}        & 42M       & 7.9G      & 83.4  \\
    % CrossFormer-B \cite{CrossFormer}   & 52M       & 9.2G      & 83.4  \\
    Focal-S \cite{Focal}       & 51M       & 9.1G       & 83.5  \\
    Shuffle-S \cite{Shuffle}   & 50M       & 8.9G      & 83.5  \\
    CSWin-S \cite{CSwin}  & 35M       & 6.9G      & 83.6  \\
    Refined-ViT-S \cite{Refiner}            & 25M       & 7.2G      & 83.6  \\
    VOLO-D1$^{\star}$ \cite{VOLO}       & 27M       & 6.8G      & 84.2  \\
    \textbf{Pale-S} (ours)       & 48M       & 9.0G      & \textbf{84.3} \\
    \textbf{Pale-S$^{\star}$} (ours)    & 48M       & 9.0G      & \textbf{85.0}  \\
    \midrule[0.5pt]
    RegNetY-16G \cite{RegNet}     & 84M       & 16.0G     & 82.9  \\ 
    ViT-B/16$^{\ddagger}$    & 86M       & 55.4G     & 77.9  \\
    PVT-L \cite{PVT}         & 61M       & 9.8G      & 81.7  \\
    DeiT-B \cite{DeiT}       & 86M       & 17.5G     & 81.8  \\
    T2T-24 \cite{T2TViT}    & 64M       & 15.0G     & 82.2  \\
    TNT-B \cite{TNT}    & 66M       & 14.1G     & 82.8  \\
    ViL-B \cite{ViL}      & 56M       & 13.4G     & 83.2  \\
    Swin-B \cite{Swin}      & 88M       & 15.4G     & 83.3  \\
    Twins-SVT-L \cite{Twins}            & 99M       & 14.8G     & 83.3  \\
    PVTv2-B5 \cite{PVTv2}   & 82M       & 11.8G     & 83.8  \\
    Focal-B \cite{Focal}   & 90M       & 16.0G     & 83.8  \\
    % PiT-B \cite{PiT}       & 74M       & 12.5G     & 84.0  \\
    Shuffle-B \cite{Shuffle}     & 88M       & 15.6G     & 84.0  \\
    % CrossFormer-L \cite{CrossFormer}    & 92M       & 16.1G     & 84.0  \\
    LV-ViT-M$^{\star}$ \cite{LV-ViT}     & 56M       & 16.0G     & 84.1  \\
    CSWin-B \cite{CSwin}        & 78M       & 15.0G     & 84.2  \\
    Refined-ViT-M \cite{Refiner}     & 55M       & 13.5G     & 84.6  \\
    VOLO-D2$^{\star}$ \cite{VOLO}       & 59M       & 14.1G     & 85.2  \\
    \textbf{Pale-B} (ours)     & 85M   & 15.6G & \textbf{84.9}  \\
    \textbf{Pale-B$^{\star}$} (ours)  & 85M & 15.6G & \textbf{85.8} \\
    \bottomrule[1pt]
\end{tabular}}
\caption{\label{EXP_CLS}Comparisons of different backbones on ImageNet-1K validation set. 
%The input size of models are set to $224\times224$, and ``${\ddagger}$'' means that the input sizes are set to $384\times384$.
All the approaches are trained and evaluated with the size of $224\times224$, except for the ViT-B$^{\ddagger}$ with size $384\times384$. The superscript ``${\star}$'' indicates employing MixToken and token labeling loss \cite{LV-ViT} during training.} 
\end{table}

% ----------------- EXP coco ---------------
\begin{table*}[t]
\centering
\resizebox{0.8\linewidth}{!}{
\begin{tabular}{l|cc|cccccc}
\toprule[1pt]
\multirow{2.2}{*}{Backbone}   & \multirow{2.2}{*}{Params}   & \multirow{2.2}{*}{FLOPs}   & \multicolumn{6}{c}{Mask R-CNN (1x)}     \\ 
%\cmidrule{4-9}
&       &       & AP$^{\text{box}}$    & AP$_{50}^{\text{box}}$     & AP$_{75}^{\text{box}}$    & AP$^{\text{mask}}$    & AP$_{50}^{\text{mask}}$    & AP$_{75}^{\text{mask}}$   \\ 
\midrule[1pt]
ResNet-50 \cite{ResNet}           & 44M        & 260G   & 38.0 & 58.6 & 41.4 & 34.4 & 55.1 & 36.7 \\
PVT-S \cite{PVT}                     & 44M        & 245G    & 40.4 & 62.9 & 43.8 & 37.8 & 60.1 & 40.3 \\
ViL-S \cite{ViL}                        & 45M        & 174G    & 41.8 & 64.1 & 45.1 & 38.5 & 61.1 & 41.4 \\
Twins-S \cite{Twins}                 & 44M        & 228G   & 42.7 & 65.6 & 46.7 & 39.6 & 62.5 & 42.6 \\
DPT-S \cite{DPT}                     & 46M        & -         & 43.1 & 65.7 & 47.2 & 39.9 & 62.9 & 43.0 \\
Swin-T \cite{Swin}                    & 48M        & 264G   & 43.7 & 66.6 & 47.6 & 39.8 & 63.3 & 42.7 \\
RegionViT-S+ \cite{RegionViT}   & 51M        & 183G    & 44.2 & 67.3 & 48.2 & 40.8 & 64.1 & 44.0 \\
Focal-T \cite{Focal}                  & 49M        & 291G    & 44.8 & 67.7 & 49.2 & 41.0 & 64.7 & 44.2 \\
PVTv2-B2 \cite{PVTv2}             & 45M        & -         & 45.3 & 67.1 & 49.6 & 41.2 & 64.2 & 44.4 \\
%Shuffle-T \cite{Shuffle}                 & 48M                      & 268G                  & -    & -    & -    & -    & -    & -    \\
CSWin-T \cite{CSwin}               & 42M        & 279G    & 46.7 & 68.6 & 51.3 & 42.2 & 65.6 & 45.4 \\
\textbf{Pale-T} (ours)  & 41M         & 306G    & \textbf{47.4}  & \textbf{69.2}  & \textbf{52.3}  & \textbf{42.7}  & \textbf{66.3}  & \textbf{46.2}   \\
\midrule[0.5pt]
%ResNet-101 \cite{ResNet}          & 63M         & 336G   & 40.4       & 61.1     & 44.2    & 36.4      & 57.7      & 38.8      \\
ResNeXt-101-32 \cite{ResNet}   & 63M         & 340G   & 41.9       & 62.5    & 45.9     & 37.5      & 59.4     & 40.2     \\
PVT-M \cite{PVT}                     & 64M         & 302G   & 42.0      & 64.4    & 45.6     & 39.0      & 61.6      & 42.1      \\
ViL-M \cite{ViL}                        & 60M         & 261G   & 43.4       & 65.9    & 47.0      & 39.7     & 62.8      & 42.1      \\
DPT-M \cite{DPT}                     & 66M         & -        & 43.8       & 66.2    & 48.3     & 40.3      & 63.1      & 43.4     \\
Twins-B \cite{Twins}                 & 76M        & 340G   & 45.1       & 67.0     & 49.4      & 41.1      & 64.1      & 44.4       \\
RegionViT-B+ \cite{RegionViT}   & 93M         & 307G   & 45.4       & 68.4    & 49.6     & 41.6      & 65.2      & 44.8     \\
%Swin-S \cite{Swin}                     & 69M        & 354G   & -           & -         & -          & -         & -          & -             \\
PVTv2-B3 \cite{PVTv2}             & 65M          & -        & 47.0       & 68.1     & 51.7     & 42.5      & 65.7      & 45.7     \\
Focal-S \cite{Focal}                  & 71M         & 401G    & 47.4      & 69.8     & 51.9       & 42.8    & 66.6      & 46.1      \\
%Shuffle-S \cite{Shuffle}                & 69M                      & 359G                    & -        & -        & -    & -    & -    & -             \\
CSWin-S \cite{CSwin}                & 54M        & 342G    & 47.9      & 70.1     & 52.6      & 43.2      & 67.1     & 46.2      \\ 
\textbf{Pale-S} (ours)                 & 68M        & 432G       & \textbf{48.4}  & \textbf{70.4}  & \textbf{53.2}  & \textbf{43.7}  & \textbf{67.7}  & \textbf{47.1}   \\
\midrule[0.5pt]
ResNeXt-101-64 \cite{ResNet}     & 101M    & 493G       & 42.8      & 63.8     & 47.3      & 38.4     & 60.6     & 41.3          \\
PVT-L \cite{PVT}                       & 81M      & 364G       & 42.9      & 65.0     & 46.6      & 39.5     & 61.9     & 42.5          \\
ViL-B \cite{ViL}                         & 76M     & 365G       & 45.1       & 67.2     & 49.3      & 41.0     & 64.3     & 44.2          \\
%Swin-B \cite{Swin}                  & -                          & -                            & -             & -             & -        & -       & -       & -             \\
Twins-L \cite{Twins}                  & 120M     & 474G       & 45.2      & 67.5     & 49.4      & 41.2     & 64.5     & 44.5          \\
PVTv2-B4 \cite{PVTv2}              & 82M      & -             & 47.5      & 68.7     & 52.0      & 42.7     & 66.1     & 46.1           \\
Focal-B \cite{Focal}                   & 110M     & 533G       & 47.8       & 70.2     & 52.5      & 43.2    & 67.3     & 46.5              \\
%Shuffle-B \cite{Shuffle}               & -                          & -                            & -             & -             & -        & -       & -       & -             \\
CSWin-B \cite{CSwin}               & 97M      & 526G       & 48.7       & 70.4     & 53.9      & 43.9    & 67.8     & 47.3          \\
\textbf{Pale-B} (ours)                & 105M     & 595G       & \textbf{49.3}    & \textbf{71.2}   & \textbf{54.1}    & \textbf{44.2}   & \textbf{68.1}   & \textbf{47.8}    \\
\bottomrule[1pt]
\end{tabular}}
\caption{\label{EXP_coco} Comparisons on COCO val2017 with Mask R-CNN framework and 1x training schedule for object detection and instance segmentation.}
\end{table*}

% ========================== Experiments  ===========================

\section{Experiments}

We first compare our Pale Transformer with the state-of-the-art Transformer backbones on ImageNet-1K \cite{ImageNet} for image classification. To further demonstrate the effectiveness and generalization of our backbone, we conduct experiments on ADE20K \cite{ADE20K} for semantic segmentation \cite{FTN, cgnet, zhang2019perspective, wu2021consensus}, and COCO \cite{COCO} for object detection \& instance segmentation. 
Finally, we dig into the design of key components of our Pale Transformer to better understand the method.

% In detail, we use AdamW \cite{adamw} optimizer with a weight decay of 0.05. The initial learning rate is set to 1e-3 and progressively decay after each iteration by a cosine schedule. The linear warmup takes up 20 epochs. We use the random horizontal flipping \cite{H_flip}, color jitter, Mixup \cite{Mixup}, CutMix \cite{cutmix} and AutoAugment \cite{AutoAugment} as data augmentation. We also adopt some common regularizations, such as Label-Smoothing \cite{Label_Smoothing} and stochastic depth \cite{Drop_Path}, but without the Exponential Moving Average \cite{EMA}. The maximal stochastic depth rate is set to 0.1, 0.3 and 0.5 for Pale-T, Pale-S and Pale-B, respectively. All the variants are trained from scratch for 300 epochs on 8 V100 GPUs with a total batch size of 1024. Only ImageNet-1k training set is used as training data. Both the training and evaluation are conducted with the input size of $224\times224$.

% (supp) All the variants are trained from scratch for 300 epochs on 8 V100 GPUs with the input size of $224\times224$ and a total batch size of 1024. During evaluation, the images are first resized to $256\times256$ and than center-cropped to $224\times224$.

\subsection{Image Classification on ImageNet-1K}
\subsubsection{Settings.} 
All the variants are trained from scratch for 300 epochs on 8 V100 GPUs with a total batch size of 1024.
%Only the ImageNet-1K training set is used as training data. 
Both the training and evaluation are conducted with the input size of $224\times224$ on ImageNet-1K dataset. 
%For fair comparisons, we follow the most settings of DeiT \cite{DeiT}, Swin \cite{Swin} and CSWin \cite{CSwin} for fair comparison. 
% we use the widely-used settings in \cite{DeiT,Swin,CSwin}.
Detailed configurations are provided in the supplementary material.

\subsubsection{Results.}
Table \ref{EXP_CLS} compares the performance of our Pale Transformer with the state-of-the-art CNNs 
and Vision Transformer backbones on ImageNet-1K validation set. 
% 和CNN比 
Compared to the advanced CNNs, our Pale variants are +3.4\%, +2.6\%, and +2.0\% better than the well-known RegNet \cite{RegNet} models, respectively, under the similar computation complexity. 
% 和其他transformer比
Meanwhile, our Pale Transformer outperforms the state-of-the-art Transformer-based backbones, and is +0.7\% higher than the most related CSWin Transformer for all variants under the similar model size and FLOPs. 
Note that LV-ViT \cite{LV-ViT} and VOLO \cite{VOLO}, using additional MixToken augmentation and token labeling loss \cite{LV-ViT} for training, seem to be on par with our approach. For fair comparisons, we use these two tricks on our Pale models, labeled by $^{\star}$ as the superscript. Pale-T$^{\star}$ achieves +0.9\% gain than LV-ViT-S$^{\star}$ with fewer computation costs. Pale-S$^{\star}$ and Pale-B$^{\star}$ achieve 85.0\% and 85.8\%, outperforming VOLO by +0.8\% and +0.6\%, respectively.

% resulting Pale-T$^{\star}$, Pale-S$^{\star}$ and Pale-B$^{\star}$.
% Note that the performance of LV-ViT \cite{LV-ViT} and VOLO \cite{VOLO} seem to be on par with our models, which use additional MixToken augmentation and token labeling loss \cite{LV-ViT} for training. 
% conduct extra experiments to equip these two tricks on our Pale models, labeled by $^{\star}$ as the superscript. 

\subsection{Semantic Segmentation on ADE20K}
\subsubsection{Settings.} 
To demonstrate the superiority of our Pale Transformer for dense prediction tasks, we conduct experiments on ADE20K with the widely-used UperNet \cite{upernet} as decoder for fair comparisons to other backbones. 
%We follow the same training settings as \cite{CSwin}. 
Detailed settings are described in the supplementary material.

%For fair comparison, we use our ImageNet-1k pretrained Pale Transformer as backbone and the widely-used UperNet \cite{upernet} as decoder, and follow the same training settings as \cite{Swin}. 
%Specially, all the models are trained for total 160k iterations with a batch size of 16. The AdamW \cite{adamw} optimizer with weight decay 0.01 is used. The initial learning rate is set to 6e-5 and decay with a polynomial scheduler after the 1500-iterations warmup. For data augmentation during training, we follow the default configurations of mmsegmentation. More details are presented in the supplementary material.
% random crop, random flipping, random rescaling (with ratio range from 0.5 to 2.0) and random photo-metric distortion 
%, which is a challenging scene parsing dataset with 150 classes

\subsubsection{Results.} 
Table \ref{EXP_ade20k} shows the comparisons of UperNet with various excellent Transformer backbones on ADE20K validation set. We report both the single-scale (SS) and multi-scale (MS) mIoU for better comparison. 
Our Pale variants are consistently superior to the state-of-the-art method by a large margin. Specifically, our Pale-T and Pale-S outperform the state-of-the-art CSWin by +1.1\% and +1.2\% SS mIoU, respectively. 
Besides, our Pale-B achieves 52.5\%/53.0\% SS/MS mIoU, surpassing the previous best by +1.3\% and +1.2\%, respectively. 
These results demonstrate the stronger context modeling capacity of our Pale Transformer for dense prediction tasks.

% ----------------- EXP ADE20K ---------------
\begin{table}[!h]
    \newcommand{\tabincell}[2]{\begin{tabular}{@{}#1@{}}#2\end{tabular}}
    \centering
    \resizebox{0.47\textwidth}{!}{
        \begin{tabular}{l|cc|ccc}
            \toprule[1pt]
            \multirow{2.2}{*}{\tabincell{l}{Backbone}}  & \multirow{2.2}{*}{\tabincell{l}{Params}}   &\multirow{2.2}{*}{\tabincell{l}{FLOPs}}      &\multirow{2}{*}{\tabincell{c}{SS \\
            mIoU}} &\multirow{2}{*}{\tabincell{c}{MS \\ mIoU}}         \\
            
              ~             & ~   & ~  & ~      & ~   \\
            \midrule[1pt]
            DeiT-S \cite{DeiT}                           & 52M       & 1099G   & -                        & 44.0   \\
            Swin-T \cite{Swin}                           & 60M       & 945G    & 44.5                    & 45.8 \\
            Focal-T \cite{Focal}                         & 62M       & 998G    & 45.8                    & 47.0 \\
            Shuffle-T \cite{Shuffle}                     & 60M       & 949G    & 46.6                    & 47.6  \\
            CrossFormer-S \cite{CrossFormer}    & 62M       & 980G    & 47.6                     & 48.4 \\
            LV-ViT-S \cite{LV-ViT}                    & 44M       & -          & 47.9                     & 48.6   \\
            CSWin-T \cite{CSwin}                      & 60M       & 959G    & 49.3                     & 50.4   \\
            \textbf{Pale-T} (ours)                               & 52M       & 996G    & \textbf{50.4}   & \textbf{51.2} \\ 
            \toprule[0.5pt]
            %ResNet-101 \cite{ResNet}  & UperNet  & 86M  & 1029G   & -  & 44.9  \\
            Swin-S \cite{Swin}                          & 81M        & 1038G   & 47.6                     & 49.5    \\
            Focal-S \cite{Focal}                        & 85M       & 1130G    & 48.0                     & 50.0    \\
            Shuffle-S \cite{Shuffle}                    & 81M       & 1044G    & 48.4                     & 49.6     \\
            VOLO-D1 \cite{VOLO}  & -  & -  & -  & 50.5 \\ 
            LV-ViT-M \cite{LV-ViT}   & 77M       & -          & 49.4                     & 50.6   \\
            CrossFormer-B \cite{CrossFormer}   & 84M       & 1090G    & 49.7                     & 50.6   \\
            CSWin-S \cite{CSwin}                     & 65M       & 1027G    & 50.0                     & 50.8      \\
            \textbf{Pale-S} (ours)                              & 80M       & 1135G     & \textbf{51.2}  & \textbf{52.2}    \\
            \toprule[0.5pt]
            Swin-B \cite{Swin}                         & 121M	 & 1188G	 & 48.1	                 & 49.7   \\
            Shuffle-B \cite{Shuffle}                   & 121M       & 1196G     & 49.0                     & 50.5    \\
            Focal-B \cite{Focal}	                    & 126M       & 1354G	 & 49.0	                 & 50.5   \\
            CrossFormer-L \cite{CrossFormer}  & 126M       & 1258M    & 50.4                     & 51.4   \\
            CSWin-B \cite{CSwin}                    & 109M      & 1222G    & 50.8                     & 51.7    \\
            LV-ViT-L \cite{LV-ViT}   & 209M       & -    & 50.9                     & 51.8   \\
            \textbf{Pale-B} (ours)                             & 119M       & 1311G     & \textbf{52.2}  & \textbf{53.0}    \\ 
            \bottomrule[1pt]
    \end{tabular}}
    \caption{\label{EXP_ade20k}Comparisons of different backbones with UperNet as decoder on ADE20K for semantic segmentation. All backbones are pretrained on ImageNet-1K with the size of $224\times224$. FLOPs are calculated with a resolution of $512\times2048$.}
\end{table}

\subsection{Object Detection and Instance Segmentation on COCO}
\subsubsection{Settings.} 
We evaluate the performance of our Pale Transformer backbone on COCO benchmark for object detection and instance segmentation, 
utilizing Mask R-CNN \cite{maskrcnn} framework under 1x schedule (12 training epochs). 
% We follow the same training strategies as \cite{Swin,CSwin}. 
Details can be found in the supplementary material.

\subsubsection{Results.} 
As shown in Table \ref{EXP_coco}, for object detection, our Pale-T, Pale-S, and Pale-B achieve 47.4, 48.4, and 49.2 box mAP for object detection, 
surpassing the previous best CSWin Transformer by +0.7, +0.5, and +0.6, respectively. 
Besides, our variants also have consistent improvement on instance segmentation, 
which are +0.5, +0.5, and +0.3 mask mAP higher than the previous best backbone.

% we compare the results of Mask R-CNN with various Transformer backbones under the 1× schedule (12 training epochs).

\subsection{Ablation Study}
We conduct ablation studies for the key designs of our Pale Transformer on image classification and downstream tasks. 
All the experiments are performed with the Tiny variant under the same training settings as mentioned above. 
We also analyze the influence of position encoding in the supplementary material.
% We adopt the same settings as mentioned above for all the ablation study experiments.

\subsubsection{Effect of Pale Size.} 
The pale sizes of four stages $\{S_1, S_2, S_3, S_4\}$ control the trade-off between the richness of contextual information and computation costs. 
As shown in Table \ref{ablation_pale_size}, increasing the pale size (from 1 to 7) can continuously improve performance across all tasks, while further up to 9 does not bring obvious and consistent improvements but more FLOPs. Therefore, we use $S_i=7, i\in\{1,2,3,4\}$ for all the tasks by default. 

%presents the effect of different pale sizes. 
%We can see that increasing the pale size (from 1 to 7) can continuously improve performance across all tasks, while is different when 

%The performance improves with the increase of pale size, especially for downstream tasks. 
%For example, as the pale size increase from 1 to 7, the Top-1 accuracy has 1.0\% gain, 
%while mIoU and box/mask mAP improve by a larger margin of 2.5\% and 1.3/1.2 respectively. 
%Thus, we used $S_i=7, i\in\{1,2,3,4\}$ for all tasks. 

% ----------------- ablation: pale size (little) ------------------
\begin{table}[t]
\newcommand{\tabincell}[2]{\begin{tabular}{@{}#1@{}}#2\end{tabular}}
\centering
\resizebox{0.98\linewidth}{!}{
\begin{tabular}{c|c|c|cc}
\toprule[1pt]
\multirow{2.1}{*}{\tabincell{c}{Pale size\\in four stages}}   & ImageNet-1K    & ADE20K    & \multicolumn{2}{c}{COCO} \\
%\cmidrule{2-11}
& Top-1 (\%)  & SS mIoU (\%)  & AP$^{\text{box}}$  & AP$^{\text{mask}}$ \\
\midrule[1pt]
\makecell[c]{1 1 1 1}   & 82.4    & 47.9    & 46.1   & 41.5  \\
\makecell[c]{3 3 3 3}   & 82.9    & 49.4    & 46.7   & 42.3  \\
\makecell[c]{5 5 5 5}   & 83.1    & 49.7    & 46.8   & 42.4  \\
\makecell[c]{\textbf{7 7 7 7}}    & \textbf{83.4}   & \textbf{50.4}   & \textbf{47.4}   & \textbf{42.7}   \\
\makecell[c]{9 9 9 9}   & 83.3    & 50.6    & 47.4   & 42.6  \\
\bottomrule[1pt]
\end{tabular}}
\caption{\label{ablation_pale_size} Ablation study for different choices of pale size. The complete table with parameters and FLOPs can be found in the supplementary material.}
\end{table}

% ----------------- ablation: attention mode (little) ------------------
\begin{table}[t]
\centering
\resizebox{0.98\linewidth}{!}{
\begin{tabular}{c|c|c|cc}
\toprule[1pt]
\multirow{2.2}{*}{Attention mode}   & ImageNet-1K    & ADE20K    & \multicolumn{2}{c}{COCO} \\
%\cmidrule{2-11}
& Top-1 (\%)  & SS mIoU (\%)  & AP$^{\text{box}}$  & AP$^{\text{mask}}$ \\
\midrule[1pt] 
Axial            & 82.4    & 47.9    & 46.1   & 41.5  \\
Cross-Shaped     & 82.8    & 49.0    & 46.6   & 42.2  \\
\midrule[0.5pt]
Pale (vanilla)   & 83.4    & 50.3    & 47.1   & 42.3  \\
Pale (sequential)    & 82.9    & 49.5    & 46.9   & 42.2  \\
Pale (parallel)    & \textbf{83.4}   & \textbf{50.4}        & \textbf{47.4}   & \textbf{42.7}   \\
\bottomrule[1pt]
\end{tabular}}
\caption{\label{ablation_attn_mode} Ablation study for different attention modes. Params and FLOPs of all the experiments are provided in the supplementary material.}
\end{table}

\subsubsection{Comparisons with Different Implementations of PS-Attention.} We compare three implementations of our PS-Attention. The vanilla PS-Attention directly conducts self-attention within the whole pale region, 
which can be approximated as two more efficient implementations, sequential and parallel. 
The sequential one computes self-attention in row and column directions alternately in consecutive blocks, while the parallel one performs row-wise and column-wise attention in parallel within each block. As shown in Table \ref{ablation_attn_mode}, the parallel PS-Attention achieves the best results on all tasks, even slightly better than the vanilla one by +0.3/0.4 box/mask mAP on COCO. We attribute this to that the excessive padding for the non-square input size in vanilla PS-Attention will result in slight performance degradation.

\subsubsection{Comparisons with other Axial-based Attentions.} 
In order to compare our PS-Attention with the most related axial-based self-attention mechanisms directly, we replace the PS-Attention of our Pale-T with the axial self-attention \cite{Axial_Deeplab} and cross-shaped window self-attention \cite{CSwin}, respectively. 
% Both of them are implemented in a parallel way for fair comparisons. 
As shown in Table \ref{ablation_attn_mode}, our PS-Attention outperforms these two mechanisms obviously.
%is clearly superior to them, 
%outperforming cross-shaped winattention by +0.6\% on ImageNet-1K, +1.4\% on ADE20K, +0.8\% bbox AP and +0.5\% mask AP on COCO.
% which treat a single row/column and multiple adjacent rows/columns as a group for attention,

% ========================== Conclusion  ===========================

\section{Conclusion}
This work presented a new effective and efficient self-attention mechanism, named Pale-Shaped self-Attention (PS-Attention), 
which performs self-attention in a pale-shaped region. 
PS-Attention can model richer contextual dependencies than the previous local self-attention mechanisms. 
In order to further improve its efficiency, we designed a parallel implementation for PS-Attention, which decomposes the self-attention within the whole pale into row-wise and column-wise attention. 
%It is also conducive to avoid excessive padding operations, 
%which is required for integral pale partition.
It is also conducive to avoiding excessive padding operations.
Based on the proposed PS-Attention, we developed a general Vision Transformer backbone, called Pale Transformer, which can achieve state-of-the-art performance on ImageNet-1K for image classification. Furthermore, our Pale Transformer is superior to the previous Vision Transformer backbones on ADE20K for semantic segmentation, and COCO for object detection \& instance segmentation.

% ========================== References ==========================

%\bibliographystyle{aaai22}
%\bibliography{aaai22}
\bibliography{aaai22}

% ========================== Appendix ==========================

\appendix

\section{Appendix}

In this appendix, we first provide the detailed experimental settings for classification, semantic segmentation, object detection, and instance segmentation, respectively. Then, we study the effect of the position encoding method of our Pale Transformer, and provide more detailed comparisons of the ablation studies in the body of our paper in terms of the model size and computation costs. Finally, the detailed derivations of computation complexity for the global self-attention and our PS-Attention are given.

% \noindent 

\section{Detailed Experimental Settings}

\subsection{Image Classification on ImageNet-1K}
We follow most of the settings in DeiT \cite{DeiT}, Swin \cite{Swin} and CSWin \cite{CSwin} for fair comparisons. In detail, we use AdamW \cite{adamw} optimizer with a weight decay of 0.05. The initial learning rate is set to 1e-3 and progressively decays after each iteration by a cosine schedule. The linear warmup takes up 20 epochs. We use the random horizontal flipping \cite{H_flip}, color jitter, Mixup \cite{Mixup}, CutMix \cite{cutmix} and AutoAugment \cite{AutoAugment} as data augmentation. We also adopt some common regularizations, such as Label-Smoothing \cite{Label_Smoothing} and stochastic depth \cite{Drop_Path}. The maximal stochastic depth rate is set to 0.1, 0.3, and 0.5 for Pale-T, Pale-S, and Pale-B, respectively. All the variants are trained from scratch for 300 epochs on 8 V100 GPUs with the input size of $224\times224$ and a total batch size of 1024. During the evaluation, the images are first resized to $256\times256$ and then center-cropped to $224\times224$.

\subsection{Semantic Segmentation on ADE20K}
We conduct experiments on the widely-used and challenging ADE20K \cite{ADE20K} scene parsing dataset, which contains 20210, 2000, and 3352 images for training, validation, and testing, respectively, with 150 fine-grained object categories. For fair comparisons, we use our ImageNet-1k pretrained Pale Transformer as backbone and UperNet as the decoder, and follow the same training settings as \cite{Swin}. Specifically, all the models are trained for total 160k iterations with a batch size of 16. The AdamW \cite{adamw} optimizer with weight decay 0.01 is used. The initial learning rate is set to 6e-5 and decay with a polynomial scheduler after the 1500-iterations warmup. The stochastic depth rate is set to 0.3, 0.3, and 0.5 for Pale-T, Pale-S, and Pale-B, respectively.
Both single-scale and multi-scale inference are reported for performance comparison. For multi-scale inference, factors vary from 0.75 to 1.75 with 0.25 as the interval. Auxiliary losses are added to the output of stage 3 of the backbone with factor 0.4, which is the same as the previous works \cite{Swin,CSwin} for fair comparisons.
For data augmentation during training, we follow the default configurations of mmsegmentation, such as random crop, random flipping, random rescaling (with ratio range from 0.5 to 2.0), and random photometric distortion.

\subsection{Object Detection \& Instance Segmentation on COCO}

We compare the performance of our Pale Transformer backbone on COCO benchmark for object detection and instance segmentation, with the typical Mask R-CNN \cite{maskrcnn} framework. We follow the same training strategies as \cite{CSwin}. In detail, we train and evaluate our Pale Transformer under the normal 1x schedule, and all the models are trained for 12 epochs on 8 GPUs with the total batch size of 16 and single-scale input (shorter size is resized to 800 and longer size is no more than 1333). We use AdamW \cite{adamw} as the optimizer with a weight decay of 0.001 for Pale-T and Pale-S and 0.05 for Pale-B. For all models, the learning rate is set to 0.0001 initially and decay at epoch 8 and 11 with a ratio of 0.1. We set the stochastic depth rate to 0.2, 0.3, and 0.5 for Pale-T, Pale-S, and Pale-B, respectively. FLOPs are compared under the input size of $1280\times800$.

% ----------------- ablation: pale size ------------------
\begin{table*}[t]
\newcommand{\tabincell}[2]{\begin{tabular}{@{}#1@{}}#2\end{tabular}}
\centering
\resizebox{0.95\linewidth}{!}{
\begin{tabular}{c|ccc|ccc|cccc}
\toprule[1pt]
\multirow{2}{*}{\tabincell{c}{Pale size\\in four stages}}   & \multicolumn{3}{c|}{ImageNet-1K}    & \multicolumn{3}{c|}{ADE20K}    & \multicolumn{4}{c}{COCO} \\
%\cmidrule{2-11}
                  & Params & FLOPs & Top-1 (\%)     & Params & FLOPs & SS mIoU (\%)     & Params  & FLOPs  & AP$^{\text{box}}$  & AP$^{\text{mask}}$ \\
\midrule[1pt]
\makecell[c]{1 1 1 1}   & 22M   & 3.8G  & 82.4         & 52M  & 929G  & 47.9                & 41M  & 253G  & 46.1  & 41.5  \\
\makecell[c]{3 3 3 3}   & 22M   & 4.1G  & 82.9         & 52M  & 950G  & 49.4              & 41M  & 269G  & 46.7  & 42.3  \\
\makecell[c]{5 5 5 5}   & 22M   & 4.4G  & 83.1         & 52M  & 972G  & 49.7              & 41M  & 283G  & 46.8  & 42.4  \\
\makecell[c]{\textbf{7 7 7 7}}    & 22M   & 4.2G  & \textbf{83.4}      & 52M  & 996G  & 50.4      & 41M  & 306G  & \textbf{47.4}  & \textbf{42.7}   \\
\makecell[c]{9 9 9 9}    & 22M   & 5.4G  & 83.3     & 52M  & 1021G  & \textbf{50.6}              & 41M  & 322G  & \textbf{47.4}  & 42.6  \\
\bottomrule[1pt]
\end{tabular}}
\caption{\label{ablation_pale_size} Ablation study for different choices of pale size.}
\end{table*}

% ----------------- ablation: attention mode ------------------
\begin{table*}[t]
\centering
\resizebox{0.95\linewidth}{!}{
\begin{tabular}{c|ccc|ccc|cccc}
\toprule[1pt]
\multirow{2}{*}{Attention mode}    & \multicolumn{3}{c|}{ImageNet-1K}    & \multicolumn{3}{c|}{ADE20K}    & \multicolumn{4}{c}{COCO} \\
%\cmidrule{2-11}
                  & Params & FLOPs & Top-1 (\%) & Params & FLOPs & SS mIoU (\%)   & Params   & FLOPs    & AP$^{\text{box}}$  & AP$^{\text{mask}}$ \\
\midrule[1pt] 
Axial           & 22M      & 3.8G      & 82.4                    & 52M  & 929G  & 47.9                     & 41M   & 253G   & 46.1  & 41.5  \\
Cross-Shaped    & 22M      & 4.2G      & 82.8                    & 52M  & 996G  & 49.0            & 41M   & 306G   & 46.6  & 42.2  \\
\midrule[0.5pt]
Pale (vanilla)     & 22M  & 5.4G  & 83.4         & 52M  & 2677G  & 50.2                     & 41M   & 668G          & 47.1    & 42.3  \\
Pale (sequential)  & 22M      & 4.2G      & 82.9                    & 52M  & 996G  & 49.5                     & 41M   & 306G   & 46.9   & 42.2  \\
\textbf{Pale (parallel)}           & 22M      & 4.2G      & \textbf{83.4}        & 52M  & 996G  & \textbf{50.4}         & 41M   & 306G   & \textbf{47.4}   & \textbf{42.7}   \\
\bottomrule[1pt]
\end{tabular}}
\caption{\label{ablation_attn_mode} Ablation study for different attention modes.}
\end{table*}

\section{Further Ablation Study}
In this section, we first provide the complete version of Table 5 and Table 6 in the body of our paper, including the parameters and FLOPs comparisons, shown in Table \ref{ablation_pale_size} and Table \ref{ablation_attn_mode}. Then, we analyze the effect of different position encoding methods for our Pale Transformer backbone.

% ----------------- ablation: position ------------------
\begin{table*}[!h]
\newcommand{\tabincell}[2]{\begin{tabular}{@{}#1@{}}#2\end{tabular}}
\centering
\resizebox{0.95\linewidth}{!}{
\begin{tabular}{c|ccc|ccc|cccc}
\toprule[1pt]
\multirow{2}{*}{\tabincell{c}{Position \\ Encoding}}   & \multicolumn{3}{c|}{ImageNet-1K}    & \multicolumn{3}{c|}{ADE20K}    & \multicolumn{4}{c}{COCO} \\
%\cmidrule{2-11}
                  & Params & FLOPs & Top-1 (\%)     & Params & FLOPs & SS mIoU (\%)     & Params  & FLOPs  & AP$^{\text{box}}$  & AP$^{\text{mask}}$ \\
\midrule[1pt]
\makecell[c]{no pos.} & 22M  & 4.2G  & 82.5    & 51M  & 996G  & 48.3     & 41M  & 306G  & 46.2  & 41.5  \\
\makecell[c]{APE}     & 22M  & 4.2G  & 82.9    & 59M  & 996G  & 49.6     & 49M  & 306G   & 46.8  & 42.2  \\ 
\makecell[c]{\textbf{CPE}}     & 22M  & 4.2G  & \textbf{83.4}      & 52M  & 996G  & \textbf{50.4}    & 41M  & 306G  & \textbf{47.4}  & \textbf{42.7}   \\
% \makecell[c]{LePE}   & 22M  & 4.2G  & 83.3       & 52M  & 996G  & 50.2     & 41M  & 306G   & 47.3  & 42.5  \\
\bottomrule[1pt]
\end{tabular}}
\caption{\label{ablation_position} Ablation study for different position encoding methods.}
\end{table*}

\subsubsection{Effect of Position Encoding.} 
The position encoding plays an important role in Transformers, as it can introduce the spatial location awareness for feature aggregation of self-attention. Here, we compare several widely-used position encoding methods, e.g., no position encoding (no pos.), absolute position encoding (APE) \cite{ViT} and conditional position encoding (CPE) \cite{CPVT}. As shown in Table \ref{ablation_position}, CPE performs best. Not using any position encoding will cause serious performance degradation, which demonstrates the effectiveness of the position encoding in Vision Transformer models.

% =========================================================================

\section{Derivations of the Computational Complexity} 

In this section, we derive the computation complexity of the global self-attention and our PS-Attention in detail.

\subsection{Computational Complexity of Global Self-Attention} 
Supposing that the size of input feature map is denoted as $h\times w\times c$. The global self-attention \cite{ViT} has three parts. Firstly, the input feature $X\in\mathcal{R}^{h \times w \times c}$ is first sent into three independent linear layers to generate query $Q\in\mathcal{R}^{h \times w \times c}$, key $K\in\mathcal{R}^{h \times w \times c}$, and value $V\in\mathcal{R}^{h \times w \times c}$, respectively. Thus, the computational complexity of the generation of $Q$, $K$, and $V$ is
\begin{equation} 
    \begin{aligned}
	& \mathcal{O}_\text{Global}^{\text{qkv}} = 3hwc^2.
    \end{aligned}
\end{equation}
Secondly, the attention map $A$ is computed by $\text{softmax}(QK^T/\sqrt{d})$. Then, the aggregated feature is obtained by the matrix multiplication between the normalized attention map $A$ and the value $V$. The computational complexity of these two processes is 
\begin{equation} 
    \begin{aligned}
	& \mathcal{O}_\text{Global}^{\text{attn}} = 2c(hw)^2.
    \end{aligned}
\end{equation}
Finally, the aggregated feature also needs to pass through a linear projection layer generally with the complexity of
\begin{equation} 
    \begin{aligned}
	& \mathcal{O}_\text{Global}^{\text{proj}} = hwc^2.
    \end{aligned}
\end{equation}
Thus, the overall computational complexity of the global self-attention is 
\begin{equation} 
    \begin{aligned}
    \label{pale:global}
	& \mathcal{O}_\text{Global} = \mathcal{O}_\text{Global}^{\text{qkv}} + \mathcal{O}_\text{Global}^{\text{attn}} + \mathcal{O}_\text{Global}^{\text{proj}} \\
	& \quad \quad \ \ \ = 4hwc^2 + 2c(hw)^2.
    \end{aligned}
\end{equation}

\subsection{Computational Complexity of Our PS-Attention} 
Similarly, given the input feature of size $h\times w\times c$ and pale size $(s_r, s_c)$, our PS-Attention(parallel) also contains three processes. Firstly, the three individual $3\times3$ separable convolutions are used to generate the query $Q\in\mathcal{R}^{h \times w \times c}$, key $K\in\mathcal{R}^{h \times w \times c}$, and value $V\in\mathcal{R}^{h \times w \times c}$, respectively, with the complexity of
\begin{equation} 
    \begin{aligned}
	& \mathcal{O}_\text{Pale}^{\text{qkv}} = 3(9hwc + hwc^2)=27hwc+3hwc^2.
    \end{aligned}
\end{equation}
Second, we decompose the self-attention within the whole pale region into row-wise and column-wise self-attention. The computational complexity of these two parallel branches are as follows
\begin{equation} 
    \begin{aligned}
	& \mathcal{O}_\text{Pale}^{\text{row}} = hw^2cs_r, \\
	& \mathcal{O}_\text{Pale}^{\text{column}} = h^2wcs_c. \\
    \end{aligned}
\end{equation}
Finally, the linear projection layer has the complexity of 
\begin{equation} 
    \begin{aligned}
	& \mathcal{O}_\text{Pale}^{\text{proj}} = hwc^2.
    \end{aligned}
\end{equation}
Therefore, the overall complexity of our parallel PS-Attention is 
\begin{equation} 
    \begin{aligned}
	& \mathcal{O}_\text{Pale} = \mathcal{O}_\text{Pale}^{\text{qkv}} + \mathcal{O}_\text{Pale}^{\text{row}} + \mathcal{O}_\text{Pale}^{\text{column}} + \mathcal{O}_\text{Pale}^{\text{proj}} \\
	& \quad \quad = 4hwc^2 + hwc(s_ch+s_rw+27).
    \end{aligned}
\end{equation}
Compared with the global self-attention, our parallel PS-Attention has lower complexity, since $2hw >> (s_ch+s_rw+27)$ always holds. 

% For example, supposing that $s_r=s_c=7$ and $h=w=m$. 

\end{document}